\PassOptionsToPackage{table}{xcolor}
\documentclass[acmsmall]{acmart}

\AtBeginDocument{%
  }

\acmJournal{TOSEM}
\setcopyright{none}
\acmDOI{}
\renewcommand\footnotetextcopyrightpermission[1]{}

\usepackage{amsmath}
\usepackage{algorithmic}
\usepackage{graphicx}
\usepackage{textcomp}
\usepackage{xcolor}
\usepackage{tikz}
\usetikzlibrary{mindmap, trees, positioning, shapes, arrows.meta, shadows}
\usepackage{adjustbox}
\usepackage[most]{tcolorbox}
\usepackage{blkarray,booktabs}
\usepackage{multirow}
\usepackage{makecell}
\usepackage{pgfplots}
\pgfplotsset{compat=1.18}
\usepackage[edges]{forest}
\usepackage{rotating}
\usepackage{comment}
\usepackage{subcaption}
\usepackage{tabularx}
\usepackage{wrapfig}

\definecolor{highlight}{HTML}{E6F0FF}

\newtcolorbox{findingbox}{
  colback=blue!5,
  colframe=black!70,
  boxrule=1pt,
  arc=2pt,
  left=6pt, right=6pt, top=6pt, bottom=6pt,
  boxsep=0pt
}

\makeatletter
\tcbset{
    myhbox/.style 2 args={%
        enhanced,
        breakable,
        colback=white,
        colframe=blue!30!black,
        attach boxed title to top left={yshift*=-\tcboxedtitleheight},
        title={#2},
        boxed title size=title,
        boxed title style={%
            sharp corners,
            rounded corners=northwest,
            colback=tcbcolframe,
            boxrule=0pt,
            coltitle=white,
            top=0.5mm, bottom=0.5mm,
        },
        underlay boxed title={%
            \path[fill=tcbcolframe] (title.south west)--(title.south east)
                to[out=0, in=180] ([xshift=3mm]title.east)--
                (title.center-|frame.east)
                [rounded corners=\kvtcb@arc] |-
                (frame.north) -| cycle;
        },
        left=2mm, right=2mm, top=1mm, bottom=1mm,
        boxsep=1mm,
        boxrule=0.6pt,
        #1
    }
}
\makeatother

\begin{document}

\title{Configuring RAG Systems: What Breaks, Where Originates, and How to Fix}

\author{Liangyu Chen}
\affiliation{%
  \institution{University of Electronic Science and Technology of China}
  \city{Chengdu}
  \country{China}}
\email{cly009@std.uestc.edu.cn}

\author{Youpeng Ma}
\affiliation{%
  \institution{University of Electronic Science and Technology of China}
  \city{Chengdu}
  \country{China}}
\email{myp@std.uestc.edu.cn}

\author{Yulong Ye}
\email{yxy382@student.bham.ac.uk}
\affiliation{%
  \institution{University of Birmingham}
  \city{Birmingham}
  \country{United Kingdom}}

\author{Zezhen Xiang}
\email{zz.xiang.work@gmail.com}
\affiliation{%
  \institution{University of Electronic Science and Technology of China}
  \city{Chengdu}
  \country{China}}

  \author{Tao Chen}
\authornote{Corresponding author.}
\affiliation{%
  \institution{University of Birmingham}
  \city{Birmingham}
  \country{United Kingdom}}
\email{t.chen@bham.ac.uk}

\renewcommand{\shortauthors}{Chen et al.}

\begin{abstract}
Retrieval-Augmented Generation (RAG) has emerged as a dominant paradigm for improving Large Language Model (LLM) by intertwining it with other components, such as a vector/graph database. Therefore, the resulting RAG systems are often highly configurable, under which a non-trivial amount of failures are related to the configurations. However, there remains a lack of systematic understanding regarding the configuration issues for RAG systems. To bridge this gap, in this paper, we conduct the first large-scale empirical study of 654 real-world configuration issues collected from 10 diverse RAG systems. Drawing on this, we provide a complete categorization of the symptoms, root causes, relevant stages, and fix patterns of configuration issues for RAG systems, as well as several new observations. Our findings not only reveal the characteristics of configuration issues, but also strongly link them to the unique pipeline stages of RAG, e.g., we found that most symptoms can be observed at the early or late stage, while the stage of the root cause might not be at the same stage as where the symptom is identified. We also provide nine actionable insights to the practitioners on testing, localizing, and fixing the configuration issues for RAG systems.
\end{abstract}



\maketitle
\fancypagestyle{reviewpagestyle}[standardpagestyle]{%
  \fancyfoot[RO,LE]{\footnotesize Manuscript submitted to ACM Trans. Softw. Eng. Methodol.}}
\fancypagestyle{reviewfirstpagestyle}[firstpagestyle]{%
  \fancyfoot[RO,LE]{\footnotesize Manuscript submitted to ACM Trans. Softw. Eng. Methodol.}}
\pagestyle{reviewpagestyle}
\thispagestyle{reviewfirstpagestyle}

\section{Introduction}
\label{sec:intro}

Large Language Models (LLMs) have witnessed unprecedented growth and evolution in recent years, demonstrating remarkable capabilities across a wide spectrum of tasks~\cite{naveed2025comprehensive,chang2024survey,zhao2023survey}.
To mitigate the issue of LLM hallucination~\cite{ji2023survey,rawte2023survey,luo2024hallucination}
and outdated knowledge~\cite{he2022rethinking,shen2023chatgpt}, Retrieval-Augmented Generation (RAG) system has emerged as a dominant paradigm~\cite{DBLP:conf/nips/LewisPPPKGKLYR020,guu2020retrieval}. Typically, as a new type of software systems, the RAG systems augment LLMs with external knowledge sources, such as documents, vector databases, and retrieval pipelines, forming the backbone of many question answering systems, developer assistants, and knowledge-based applications~\cite{DBLP:journals/corr/abs-2312-10997, arslan2024survey, fan2024survey}.


To ensure flexibility, RAG systems are designed to be highly configurable~\cite{csakar2025maximizing, barker2025faster, ray2025metis, 11330856, gao2024modular}---a key property of many types of software systems. For example, changing the retrieval stride length will significantly affect the quality of the generated text in RAG systems~\cite{li2025enhancing}. However, configuration has long been recognized as a critical source of reliability issues in traditional configurable software systems~\cite{liu2024rethinking,10.1145/3243734.3243794,10.1145/2517349.2522727,han2016empirical,han2018perflearner,siegmund2015performance,he2020cp,he2019tuning,ma2025faster,xu2015hey}, RAG systems are of no exception. Yet, the unique complex stages in the pipeline of RAG systems (e.g., indexing, retrieval, and generation) make their configuration even more error-prone~\cite{barker2025faster, 11330856, ray2025metis}. For example, it has been shown that configuration issues of RAG systems (including both configuration bugs and misconfiguration) can cause a latency performance degradation of $10\times$, leading to a two orders of magnitude cost~\cite{11330856}. Another example showcases the complex interaction of configuration options among the RAG stages: a subtle change in the indexing strategy can unexpectedly invalidate assumptions in the retrieval or generation stages, causing significant errors to propagate silently across the pipeline, and hence a system failure. Therefore, understanding the configuration issues of RAG systems and how they have been fixed is in high demand.

Despite the importance, our current knowledge of the configuration issues within RAG systems remains rather limited. While there are prior empirical studies of RAG systems, they focus on two different aspects: either investigates how configuration choices influence system performance~\cite{barker2025faster,ray2025metis,11330856}, typically from a configuration tuning and hyperparameter optimization perspective; or examines failure cases and practical challenges in RAG systems through empirical analyses of observed bugs in general, which are not configuration-related~\cite{DBLP:conf/fllm/AgrawalKA024, DBLP:journals/corr/abs-2510-13975, DBLP:conf/cain/BarnettKTB024}. There has been no systematic study of the configuration issues as the root of RAG systems' failures, particularly with respect to the RAG pipeline.

To bridge this gap, in this paper, we conduct a large-scale empirical study on RAG configuration issues, covering 654 issues and 345 pull requests from 10 RAG systems on GitHub over six person-months. We make several contributions:





\begin{itemize}
    \item \textbf{A Categorization:} We summarize a comprehensive categorization for configuration issues in RAG systems with respect to its pipeline stages, including five types of symptoms, nine root causes, and six fix patterns.
    
    
    
    \item \textbf{Empirical Findings:} Our empirical study reveals ten previously unknown findings, the most notable ones are:

    \begin{itemize}
        \item Core pipeline fault, in which failures occurred at a pipeline stage with exceptions, is the predominant symptom of configuration issues in RAG systems.
        \item For configuration bugs, invocation of external libraries/systems is the most common cause; while resource provision that does not fulfil the needs is the most common reason for user misconfiguration.
        \item Most of the symptoms occurred at the initialization, indexing and generation stage of RAG systems; while root causes mostly lie in the early stages.
        \item The stage where the symptom is observed might not necessarily correspond to the stage of the root cause---as observed in one third of the cases. 
        \item When the symptom and root cause deviate at more distant stages (e.g., from initialization to the last stage of generation), the resolving cost (time) of real configuration bugs can increase by up to $2.83\times$. 
        
        \item Patching the logic and algorithm is the most common fix pattern for configuration bugs.
    \end{itemize}

    
    
    
    \item \textbf{Insights:} Based on these results, we distil several insights about configuration issues for RAG systems on their testing/detection, localization, and fixing.



\end{itemize}


All the data and code can be accessed at our repository: \textcolor{blue}{\texttt{\url{https://github.com/ideas-labo/rag-config-bugs}}}.

This paper is organized as follows. 
\autoref{sec:background} introduces the background. 
\autoref{sec:method} presents our methodology. 
\autoref{sec:results} delineates the categorization of configuration-related issues and the findings. 
\autoref{sec:insights} discusses actionable insights. 
\autoref{sec:threats} addresses threats to validity.
\autoref{sec:related_work} reviews related work. 
Finally, \autoref{sec:conclusion} concludes the paper.



\section{Background}
\label{sec:background}

    \subsection{RAG Pipeline}

\begin{figure}[t!]
    \centering
    \includegraphics[width=0.60\linewidth]{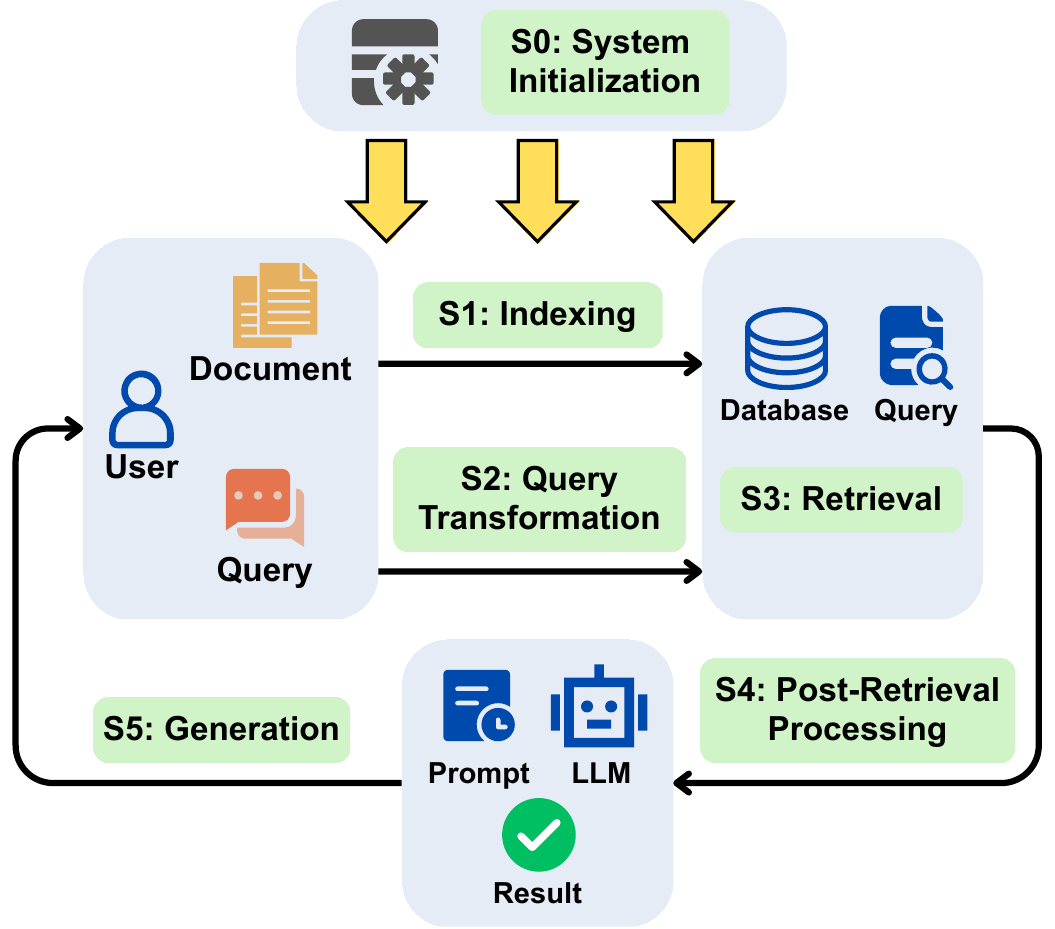}
    \caption{Workflow of RAG with different pipeline stages.}
    \label{fig:rag_workflow}
\end{figure}

As shown in Figure~\ref{fig:rag_workflow}, RAG systems operate in their unique pipeline to work with LLM. Specifically, given a task (e.g., question answering), there are several key stages~\cite{DBLP:journals/corr/abs-2312-10997}:


\begin{enumerate}
    \item[\textbf{S0:}] \textbf{Initialization} prepares, validates the environment, and fulfils the resources needed to boot the system.
   \item[\textbf{S1:}]  \textbf{Indexing} transforms raw data into a searchable knowledge base via chunking and embedding.
   \item[\textbf{S2:}]  \textbf{Query Transformation} rewrites or decomposes user inputs to improve retrieval capability. 
   \item[\textbf{S3:}]  \textbf{Retrieval} fetches contexts from vector/graph database.
   \item[\textbf{S4:}]  \textbf{Post-Retrieval Processing} refines results through re-ranking and compression. 
   \item[\textbf{S5:}]  \textbf{Generation} synthesizes the final prompt with contextual information for the LLM(s) to process.
\end{enumerate}


Each stage comes with its own configuration options. The final output serves as the answer to different tasks. 




\subsection{Configuration Issues for RAG Systems}

\begin{table}[t!]
\centering
\caption{A real-world configuration issue from RAG system \textsc{RAGFlow}.}
\label{tab:rag_config_example}

\begin{tabularx}{0.82\linewidth}{|X|}
\toprule



\textbf{ID $\#7790$:} Potential Issue in \texttt{naive\_merge\_docx} Function \\

\midrule
\textbf{Description (excerpt from the issue report):} \\
{...This can result in the first chunk being larger than intended, as there's no proper size control for the initial chunk...}  \\
{...The function initializes \texttt{tk\_nums = [0]} and uses \texttt{tk\_nums[-1] > chunk\_token\_num} as a condition for creating new chunks... Since $0 < \text{chunk\_token\_num}$ (default 128), the first section is always added to the first chunk...} \\
\bottomrule
\end{tabularx}
\end{table}

In this paper, we define a \textit{configuration issue} in RAG systems as a defect that satisfies two conditions: 

\begin{itemize}
    \item The issue must manifest as an observable system-level problem: functional defects, e.g., execution failures, crashes, or core pipeline breakdowns, and performance-related degradations, e.g., latency or answer quality.
    \item It must be directly related to configuration.
\end{itemize}


From the example in ~\autoref{tab:rag_config_example}, we see that the reported issue exhibits a system-level problem and is relevant to the configuration of the RAG system. Clearly, it indicates an observable system-level problem, i.e., a degradation in document chunking quality during indexing, with clear statements confirming that the problem is directly related to the configuration option \texttt{chunk\_token\_num}.

In general, configuration issues can be broadly distinguished by two types depending on their root causes~\cite{he2020cp,10.1145/2517349.2522727}, which are both considered in this work:


\begin{itemize}
    \item \textbf{Configuration bugs:} Here, the issue is caused by system defects, where the system fails to correctly handle specific configuration combinations due to flawed adaptation logic, incorrect assumptions, or incomplete validation.
    \item \textbf{Misconfigurations:} The issues are mainly caused by incompatible or improper configuration settings (e.g., model choices and retrieval strategies) made by the users. 
\end{itemize}

\section{Empirical Methodology}
\label{sec:method}

\begin{figure}[t!]
    \centering
    \includegraphics[width=0.70\linewidth]{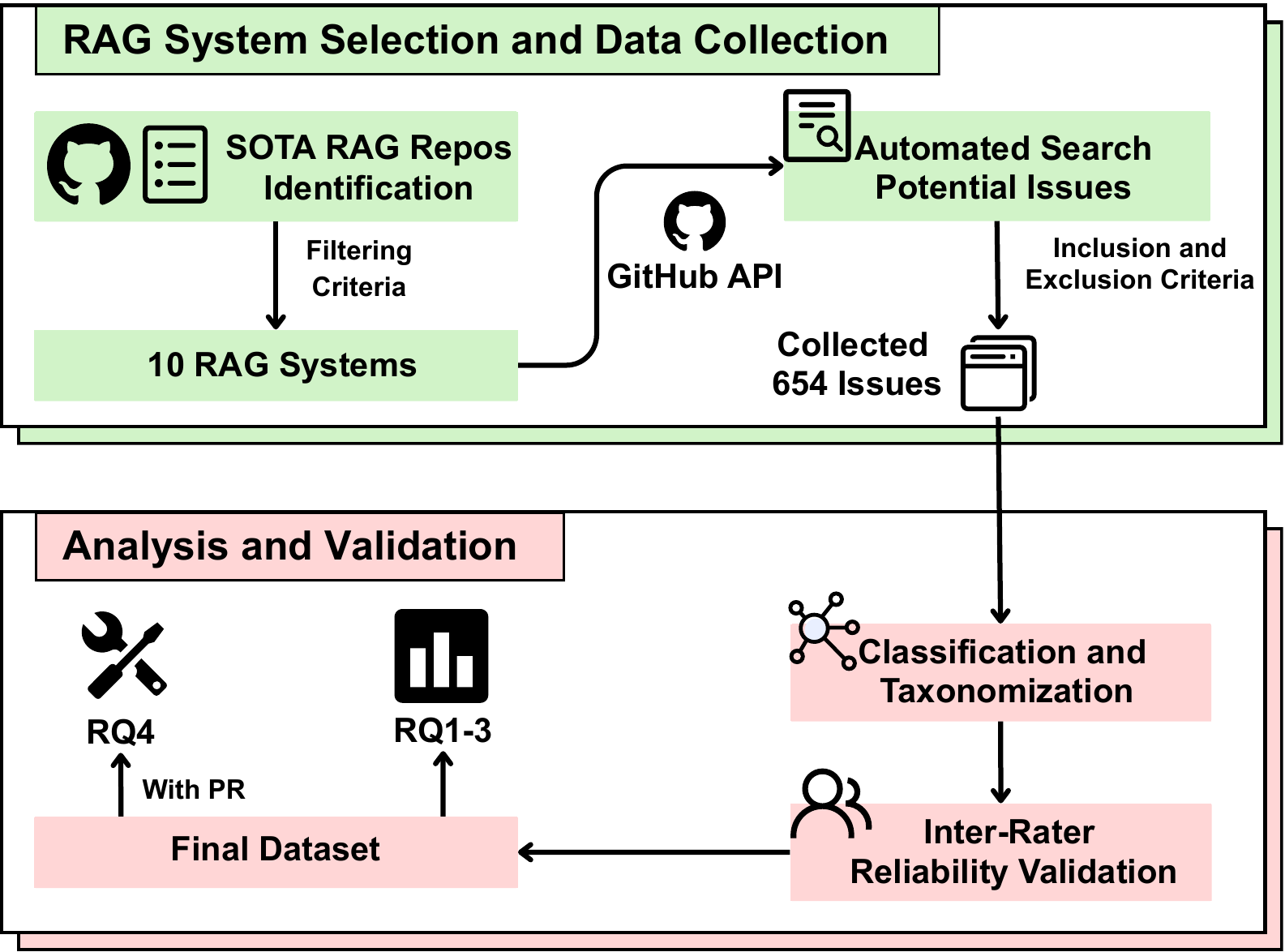}
    \caption{The empirical study methodology.}
    \label{fig:method3}
\end{figure}

\autoref{fig:method3} shows our methodology. As can be seen, we select representative RAG systems from GitHub, followed by rigorous extraction and analysis of the issues. These then allow us to construct a categorization of configuration issues for RAG systems, together with a rich set of findings. All of those are systematically validated for ensuring inter-rater reliability.

\subsection{Research Questions}
\label{sec:rq}


We seek to answer the following research questions (RQs):

\begin{itemize}
    \item \textbf{RQ1:} What common symptoms do configuration issues exhibit in RAG systems?
    \item \textbf{RQ2:} What commonly causes the configuration issues? 
    \item \textbf{RQ3:} How do the deviated stages between symptom and root cause affect the resolution of configuration issues? 
    \item \textbf{RQ4:} How are the actual configuration bugs often fixed?
    
\end{itemize}

\textbf{RQ1} and \textbf{RQ2} summarize the general characteristics of configuration issues for RAG systems. \textbf{RQ3} is derived from the fact that the stage of the exhibited symptom of a configuration issue might deviate from that of its root cause (e.g., a symptom from the \textit{indexing} stage might be caused by bugs in the \textit{generation} stage), hence we aim to study how such a phenomenon might affect its bug fixing. \textbf{RQ4} investigates the patterns of actual configuration bug fixing, if any.

Notably, we centre around the above aspects because RAG developers typically reasoned about configuration issues from three complementary pillars: (1) how the issue manifests in practice; (2) where the issue originates; and (3) how the issue is eventually resolved. Thus, understanding them enables us to analyze issue propagation across the RAG pipeline and derive actionable guidance for debugging and maintenance.

\subsection{RAG System Selection}

To identify the RAG system, we use several keywords to search over GitHub, i.e., \textit{``RAG''} or \textit{``Retrieval-Augmented Generation''}. This returns more than a thousand projects. We then filter the systems using the criteria below:



\begin{itemize}

     \item \textbf{Configuration Accessibility}: The repository contains user-adjustable configuration artefacts, such as configuration files, environment variables, or documented options. 
    
    \item \textbf{Popularity}: To ensure sufficient issue discussion and relevance, we remove repositories with less than $2,000$ stars\footnote{This was chosen pragmatically as a general measurement of popularity.}---a proxy for both community adoption and the likelihood of reporting configuration-related issues. 
   
    \item \textbf{Issue Richness}: Repositories with fewer than 20 issues are removed, as there is little information to be analyzed.

    

\end{itemize}

\begin{table}[t]
\centering
\caption{Statistics of the selected RAG systems.}
\label{tab:rag_repos}
\begin{adjustbox}{max width=0.85\linewidth,center}
\begin{tabular}{llll|rr}
\toprule
\multirow{2}{*}{\textbf{RAG System}} & \multirow{2}{*}{\textbf{Stars}} & \multicolumn{2}{c|}{\textbf{Retrieved Issues}} & \multicolumn{2}{c}{\textbf{Filtered Issues}} \\
 & & Total & Closed & Total & with PR \\
\midrule

\rowcolors{1}{white}{gray!15}

\textsc{RAGFlow}           & 61.1k & 1,417 & 685 & 217 & 161 \\
\textsc{DB-GPT}           & 17.0k & 812   & 559 & 120 & 76  \\
\textsc{FastGPT}          & 25.3k & 494   & 412 & 74  & 19  \\
\textsc{GraphRAG}         & 26.9k & 195   & 152 & 64  & 28  \\
\textsc{LightRAG}         & 18.9k & 214   & 197 & 63  & 22  \\
\textsc{Obsidian-copilot} & 5.0k  & 564   & 353 & 35  & 15  \\
\textsc{Private-GPT}       & 56.4k & 255   & 225 & 31  & 3   \\
\textsc{Quivr}            & 38.0k & 283   & 282 & 22  & 8   \\
\textsc{DocsGPT}          & 16.9k & 332   & 318 & 21  & 10  \\
\textsc{HippoRAG}         & 2.6k  & 86    & 75  & 7  & 3   \\

\midrule
\rowcolor{white}
\textbf{Total} & -- & 4,652 & 3,258 & 654 & 345 \\
\bottomrule
\end{tabular}
\end{adjustbox}
\end{table}

After applying the above criteria, we identify several tens of repositories of the RAG systems\footnote{As of 1st Sep, 2025.}. To ensure both representativeness and realism, we select the top 10 repositories ranked by the number of GitHub stars among those eligible. 

As can be seen from~\autoref{tab:rag_repos}, the RAG systems are of distinct domains, including general end-to-end RAG platforms\footnote{Some of these can be used for building other types of LLM systems, but we explicitly extract the RAG-related issues only.} (e.g., \textsc{FastGPT}), research prototypes (e.g. \textsc{HippoRAG}), and domain specific RAG system (e.g. \textsc{DocsGPT}). Further, the selected RAG systems cover the major forms of popular open-source RAG software, including end-to-end platforms, developer-oriented frameworks, research prototypes, and user-facing or domain-specific applications. They also incorporate heterogeneous RAG components and external services, such as LLMs, embedding models, vector databases, retrievers, rerankers, and data sources. This diversity provides broad coverage of the architectures, application scenarios, and configuration surfaces found in popular and active open-source RAG systems, making the selected repositories representative within the scope targeted by our study.




\subsection{Data Collection and Filtering}
\label{sec:collection_filtering}

We collected issues from the repositories of the 10 RAG systems (till 1st August 2025) using a two-step process: automated retrieval and then manual validation by the authors.

\subsubsection{Automatic Search}
To comprehensively identify configuration issues under a repository of RAG system, we constructed a set of search queries by combining two primary dimensions: commonly used configuration terms in software systems (e.g., \texttt{config}, \texttt{setting}, \texttt{parameter}, \texttt{env}) and {RAG-specific terms} corresponding to core pipeline components (e.g., \texttt{embedding}, \texttt{vectorstore}, \texttt{retriever}, \texttt{prompt}). Each pair of the above terms is used in isolation; the resulting issues are then merged with redundancy removal.


We tailored our search scope based on the volume of issues and the effectiveness of labels in each repository: 
\begin{itemize}

    \item  \textbf{Labeled Search}: For projects with a large volume of issues and well-maintained labeling systems (e.g., \textsc{LightRAG}), we combine the keywords with the project-specific label (e.g., \texttt{keyword AND label:bug}) to locate related issues while discarding unrelated discussions. 
        
    \item \textbf{Keyword-Only Search}: For projects that did not maintain a meaningful labeling protocol (or those returned fewer than 20 issues), we conducted keyword-based searches paired with manual screening. 
    
\end{itemize}



As from~\autoref{tab:rag_repos}, this process yields a preliminary dataset of {4,652} candidate issues. From this pool, we focus on the {3,258} closed issues due to the needs for \textbf{RQ3} and \textbf{RQ4}.

\subsubsection{Manual Filtering}
Since the retrieved issues might contain noise, e.g., some of them could be usage questions or feature requests. To mitigate this, we conducted a rigorous manual inspection to identify genuine configuration-related issues. We say that an issue should be included in the final dataset if and only if it meets all the criteria below:

\begin{itemize}
    \item \textbf{Validity}: The issue must describe a confirmed system defect (e.g., crash, functional error, performance degradation) or user misconfiguration (e.g., parameter/option misunderstanding, usage errors, or documentation-induced mistakes), excluding feature requests and inquiries.
    
    \item \textbf{Relevance}: The issue’s trigger, manifestation, or expected resolution must be logically attributable to at least one user-adjustable configuration option of RAG, such as those specified in configuration files (e.g., \texttt{YAML} or \texttt{TOML}), environment variables, or API arguments. 
    
    
    \item \textbf{Traceability}: Only {closed} issues are considered, and sufficient evidence (e.g., linked pull requests, reproduction steps, or conclusive developer discussions) has to be available to reasonably infer the root cause of the issues.
\end{itemize}

Applying these criteria resulted in {654} configuration issues for RAG systems in total, which we use to address \textbf{RQ1}--\textbf{RQ3}. Among them, {345} issues explicitly link to pull requests (PR), enabling a deeper analysis of code-level fix patterns for \textbf{RQ4}.



\subsection{Analysis and Validation Procedure}
\label{sec:analysis}

To systematically analyze configuration issues in RAG systems, we carefully read the issues' description, comments/discussions, and pull requests (if any), which provide rich information on the symptoms (\textbf{RQ1}), root causes (\textbf{RQ2}), and fix patterns (\textbf{RQ4}), following iterative open-coding~\cite{10.1109/ESEM.2011.36,https://doi.org/10.1111/add.13314}.

\subsubsection{Iterative Categorization}


Four authors, who are experienced researchers of RAG systems, lead the categorization. The analysis starts with an exploratory open-coding phase. We randomly sample an initial 10\% issues (stratified across systems) in an open-coded manner without predefined categories. Based on recurring patterns, we propose preliminary categories (and sub-categories as needed) for \textbf{RQ1}, \textbf{RQ2}, and \textbf{RQ4}. In the first iteration, the full dataset is then labeled, with iterative refinements (merging, splitting, redefining categories); we stop at the level of categories where the next level of sub-categories resulted in highly biased labeling---over 90\% of the issues are condensed to one sub-category---or no finer-grained common phenomenon across the RAG systems can be observed. All authors need to agree upon a final categorization.

\subsubsection{Inter-Rater Reliability Validation}


With the final categories, each author then independently conducts the second iteration, aiming to correct the labels of issues using those categories. Note that in some cases, an issue might involve multiple categories/sub-categories, in which case we label it with the most fundamental/crucial one, as agreed by all raters. We also seek to assign the issues to the relevant RAG stages for \textbf{RQ3}. To measure inter-rater reliability, we use Cohen's $\kappa$~\cite{landis1977measurement}. Through multiple iterations of relabeling, agreeing, and consulting external experts, we seek to reach $\kappa\geq0.7$ for all levels of the categories and stages\footnote{0.7 is a pragmatic ``good'' agreement in software engineering ~\cite{9772253,cheriyan2021towards}.}~\cite{9772253,cheriyan2021towards}.

As summarized in Table~\ref{tab:kappa}, we achieve $\kappa\geq0.7$ at all levels, including all the categories and identification of the RAG stage, with three iterations. In particular, we have perfect agreement for the main category of symptoms and root causes.


\subsubsection{Unweighted vs. Weighted Percentage}

To provide overall statistics, we report on both the unweighted and weighted proportions of issues per category/sub-category over all systems, i.e., the former is robust to the diverse sample sizes of the RAG systems, while the latter is sensitive to the sizes.

\begin{table}[!ht]
\caption{Inter-rater agreements (Cohen's $\kappa$) recorded.}
\label{tab:kappa}
\begin{adjustbox}{max width=0.70\linewidth,center}
  \centering
\begin{tabular}{lccc}
\toprule
 & \textbf{2nd Iteration} & \textbf{3rd Iteration} & \textbf{4th Iteration} \\
\midrule
Symptom category & 0.78 & 1.00 & 1.00 \\
Symptom sub-category & 0.61 & 0.73 & 0.73 \\
Symptom stage & 0.89 & 0.89 & 0.89 \\
\midrule
Root cause category & 0.68 & 1.00 & 1.00 \\
Root cause sub-category & 0.53 & 0.69 & 0.80 \\
Root cause stage & 0.62 & 0.74 & 0.74 \\
\midrule
Fix pattern & 0.63 & 0.82 & 0.82 \\
\bottomrule
\end{tabular}
\end{adjustbox}
\end{table}

\section{Results}
\label{sec:results}




\subsection{Categorization of RAG Configuration Issues}
\label{sec:taxonomy}


\begin{figure*}[t!]
    \centering
    \includegraphics[width=\linewidth]{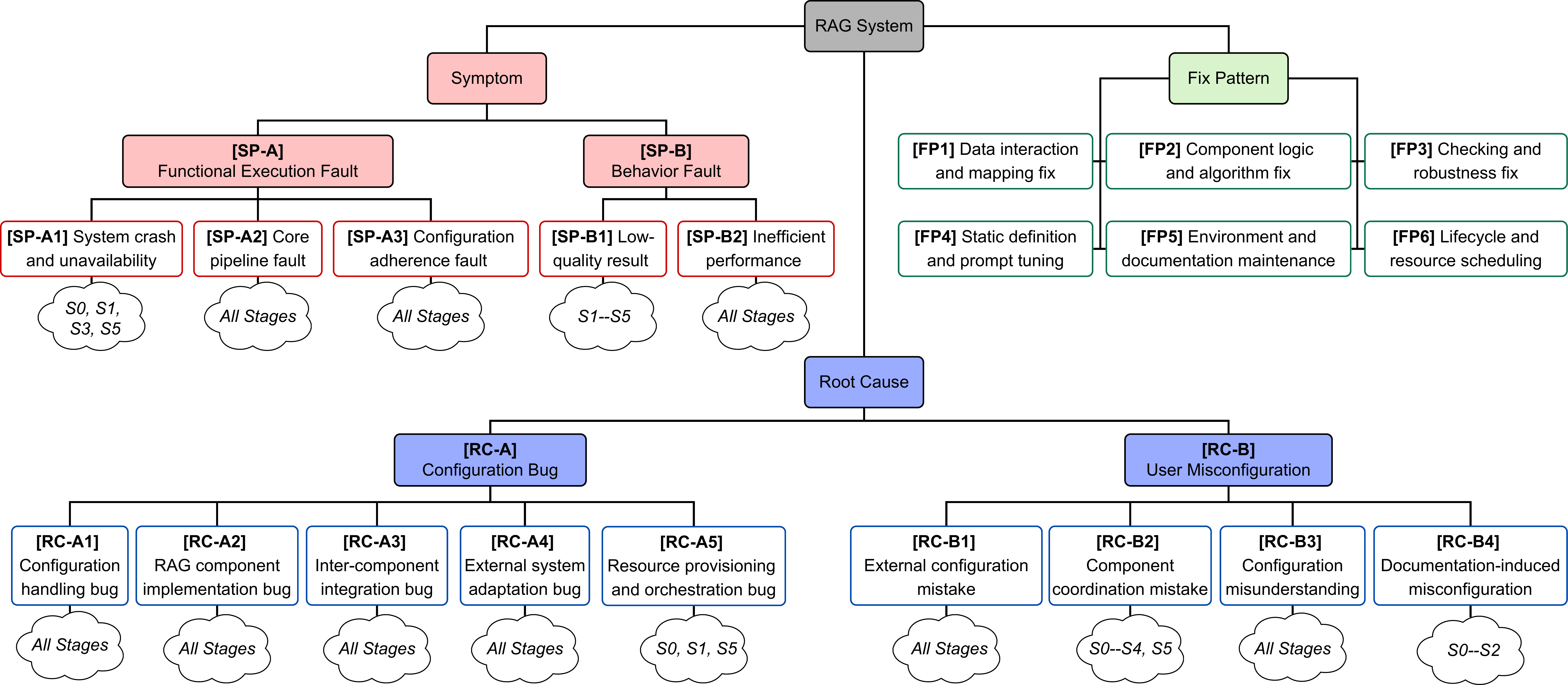}
    \caption{Categorization for configuration issues of RAG systems.}
    \label{fig:taxonomy}
\end{figure*}

Unlike others, RAG systems operate in a multi-stage semantic pipeline that integrates heterogeneous components, which have been captured by our categorization in Figure \ref{fig:taxonomy}. It is worth noting that our categorization contains both general software-failure manifestations and RAG-specific refinements. High-level symptoms such as crashes or performance degradation may also occur in conventional software systems. However, their finer-grained manifestations in our dataset are tied to the functional components of RAG systems. In particular, \textit{core pipeline faults} are distinguished according to the RAG stage in which the affected functionality resides, while \textit{external system adaptation bugs} capture incompatibilities involving LLMs, embedding models, vector databases, external data sources or applications, and rerankers. These component-specific categories are further connected to the symptom and root cause stages, providing a stage-aware characterization that is not captured by generic configuration bug categories.



\subsubsection{Symptoms}
\label{sec:taxonomy_symptom}

The Symptom dimension characterizes how configuration issues become externally observable from a user perspective. We divide the observable symptoms into two categories, each of which has different sub-categories.

\textbf{Functional Execution Fault.} This corresponds to cases where the system or a specific pipeline stage fails to fulfil the intended needs, including three sub-categories:

\begin{itemize}
    \item \textbf{System crash and unavailability} mean the system terminates unexpectedly without explicit exceptions or hangs indefinitely. The RAG system becomes entirely non-responsive without reasons, preventing any further interaction: e.g., issue $\#822$ for \textsc{Obsidian-copilot}.
    \item \textbf{Core pipeline fault} denotes that the system remains functional, but fails to fully complete a specific stage of the pipeline with explicit exceptions, causing the system to stop: e.g., issue $\#1642$ for \textsc{Graphrag}.
    \item \textbf{Configuration adherence fault} means that the system has good quality/performance, but violates the user’s configuration intent: e.g., issue $\#2636$ for \textsc{DB-GPT}.
\end{itemize}


\textbf{Behavior Fault.} Here, the RAG system suffers severe behavior-level degradations, where the system delivers \textbf{low-quality results}, i.e., the system does generate answers, but they are factually incorrect/incomplete; or exhibits \textbf{poor performance}: a good quality outcome with abnormal resource consumption or excessive latency. For example, issue $\#1321$ for \textsc{LightRAG} and issue $\#7518$ for \textsc{RAGFlow}.


\subsubsection{Root Cause}

Our results indicate that root causes of the configuration issues for RAG systems can be divided into two categories, each of which contains several sub-categories. 

\textbf{Configuration Bug.} The configuration issues whose root causes lie in the system’s code. There are five sub-categories:

\begin{itemize}
    \item \textbf{Configuration handling bug} arises when configuration options are incorrectly parsed, transmitted, or bound to execution logic, e.g., issue $\#1313$ for \textsc{DocsGPT}.
    \item \textbf{RAG component implementation bug} originates from flawed internal logic at core RAG data processing (such as chunking, retrieval strategy, graph clustering, etc.) related to configuration, e.g., issue $\#436$ for \textsc{RAGFlow}.
    \item \textbf{Inter-component integration bug} occurs at component boundaries, where mismatches between upstream outputs and downstream input assumptions violate implicit data contracts within the RAG pipeline under specific configurations. For example, issue $\#1642$ for \textsc{GraphRAG}.
    \item \textbf{External system adaptation bug} stems from incompatible configuration between the RAG system and external artefacts such as {LLMs, databases, embedding models, or re-rankers}: issue $\#990$ for \textsc{Private-GPT}.
    \item \textbf{Resource provisioning and orchestration bug} reflects failures in allocating, scheduling, or utilizing underlying computational resources for fulfilling a given configuration. For example, issue $\#1288$ for \textsc{RAGFlow}.
\end{itemize}


\textbf{User Misconfiguration.} Unlike configuration bugs, user misconfiguration mainly originates from users' misunderstanding of the RAG system's configuration options:


\begin{itemize}
    \item \textbf{External configuration mistake} occurs when users select or configure external artefacts---such as LLMs or databases---that are incapable of supporting the intended workload. For example, issue $\#1567$ for \textsc{LightRAG}.
    \item \textbf{Component coordination mistake} arises when users set the configuration to combine incompatible internal RAG components, violating functional dependencies between pipeline stages. For example, issue $\#614$ for \textsc{FastGPT}.
    \item \textbf{Configuration misunderstanding} reflects misunderstandings of configuration semantics, constraints, or deprecated options, leading to behaviors that diverge from user intents despite technically valid inputs. For example, issue $\#144$ for \textsc{Obsidian-copilot}.
    \item \textbf{Documentation-induced misconfiguration} captures cases where users are misled by incorrect, outdated, or ambiguous documentation, resulting in configurations that follow official guidance but fail in practice. For example, issue $\#8556$ for \textsc{RAGFlow}.
\end{itemize}


\subsubsection{Stage Labeling}
\label{sec:taxonomy_stage}

We also study how configuration issues are localized within and propagated through the RAG pipeline with respect to the symptoms and root cause\footnote{We found that most symptoms/root causes can correspond to one stage only. For the rare cases of symptoms/root causes that are related to multiple stages, we use the one that is the most relevant, as agreed by all authors.}:

\begin{itemize}
    
    \item \textbf{Symptom Stage:} The stage where the abnormal behavior first becomes observable to the user. Crucially, this measures failure observability rather than the exact moment an internal pipeline data flow might first deviate silently.
    

    \item \textbf{Root Cause Stage:} This pinpoints where the configuration issue originates. For configuration bugs, it is the stage of the flawed code logic; for {user misconfiguration}, it is the stage where users set improper configuration options. We infer the root-cause stage by jointly analyzing issue content and developer discussions. For issues with pull requests, we examine the modified code and the comments to identify the originating pipeline stage. 
    
    
\end{itemize}

We found that some symptoms/root causes are linked to all stages, while others are only related to certain stages.




\subsubsection{Fix Pattern}

We extract six fix patterns for configuration issues in RAG systems based on 345 issues with pull requests:


\begin{itemize}
    \item \textbf{Data interaction and mapping:} This pattern involves ensuring configuration-related data mapping and interface alignment between disparate RAG components or external APIs (e.g., key renaming, context propagation).
    
    \item \textbf{Component logic and algorithm:} This involves modifying the core business logic or internal algorithms (e.g., rewriting chunking flows) to better handle configurations. 
    
    \item \textbf{Checking and robustness:} This pattern focuses on increasing system stability by adding error checking/warning code, e.g., null-pointer checkers, data type validation, and improved error-handling logic to prevent crashes under invalid or incompatible configurations.
    
    \item \textbf{Static definition and prompt tuning:} These fixes involve tuning static constants, default values, or prompt templates. Developers often use this to refine hard-coded defaults or update prompt constraints within the \textit{generation} stage to align with the intents of the configurations.
    
    \item \textbf{Environment and documentation maintenance:} This covers configuration-related but non-code interventions, such as fixing external dependencies, updating deployment (e.g., Docker), or correcting misleading documentation to ensure correct RAG system deployment.
    
    \item \textbf{Lifecycle and resource scheduling} This involves governing configuration-related resources for RAG's lifecycle, including concurrency and consistency. Typical fixes include singleton enforcement, implementing cache invalidation, and optimizing pooling for external databases.
\end{itemize}

\subsection{RQ1: Symptoms of Configuration Issues}
\label{sec:rq1}

\begin{table}[t!]
\centering
\caption{Symptoms proportion by systems/categories---\% per system (count). The highest \% is highlighted in \colorbox[HTML]{E6F0FF}{blue}.}
\label{tab:symptom_distribution}

\setlength{\tabcolsep}{1mm}
\begin{adjustbox}{max width=0.90\linewidth,center}
\begin{tabular}{llll|ll|r}
\toprule
\textbf{System} & \textbf{SP-A1} & \textbf{SP-A2} & \textbf{SP-A3} & \textbf{SP-B1} & \textbf{SP-B2} & \textbf{Total} \\
\midrule

\rowcolors{1}{white}{gray!15}

\textsc{RAGFlow}          & 8\% (17)  & \cellcolor{highlight}63\% (137) & 17\% (37) & 9\% (19)  & 3\% (7)   & 217 \\
\textsc{DB-GPT}           & 11\% (13) & \cellcolor{highlight}66\% (79)  & 9\% (11)  & 13\% (16) & 1\% (1)   & 120 \\
\textsc{FastGPT}          & 4\% (3)   & \cellcolor{highlight}55\% (41) & 18\% (13) & 20\% (15) & 3\% (2)   & 74  \\
\textsc{GraphRAG}         & 6\% (4)   & \cellcolor{highlight}59\% (38)  & 23\% (15) & 9\% (6)   & 2\% (1)   & 64  \\
\textsc{LightRAG}         & 11\% (7)  & \cellcolor{highlight}63\% (40)  & 6\% (4)   & 14\% (9)  & 5\% (3)   & 63  \\
\textsc{Obsidian-copilot} & 6\% (2)   & \cellcolor{highlight}77\% (27)  & 6\% (2)   & 11\% (4)  & 0\% (0)        & 35  \\
\textsc{Private-GPT}      & 6\% (2)   & \cellcolor{highlight}74\% (23)  & 3\% (1)   & 6\% (2)   & 10\% (3)  & 31  \\
\textsc{Quivr}            & 5\% (1)   & \cellcolor{highlight}86\% (19)  & 5\% (1)   & 5\% (1)   & 0\% (0)        & 22  \\
\textsc{DocsGPT}          & 14\% (3)  & \cellcolor{highlight}62\% (13)  & 14\% (3)  & 0\% (0)   & 10\% (2)  & 21  \\
\textsc{HippoRAG}         & 29\% (2)  & \cellcolor{highlight}57\% (4)   & 14\% (1)  & 0\% (0)   & 0\% (0)        & 7   \\

\midrule
\rowcolor{white}
\textbf{Weighted} & 8\% (54) & \cellcolor{highlight}64\% (421) & 14\% (88) & 11\% (72) & 3\% (19) & 654 \\
\textbf{Unweighted} & 10\% (54) & \cellcolor{highlight}66\% (421) & 12\% (88) & 9\% (72) & 3\% (19) & 654 \\
\bottomrule
\end{tabular}
\end{adjustbox}
\end{table}







\subsubsection{Commonality of Symptoms}

As from \autoref{tab:symptom_distribution}, the observed \textit{functional execution fault} (86\%/88\% for SP-A1, SP-A2, and SP-A3) is predominantly more common than \textit{behavior fault} (14\%/12\% for SP-B1 and SP-B2) on both weighted and unweighted percentages. These results indicate that configuration issues in RAG systems primarily act as execution blockers rather than behaviors influencers. Within the former, for all RAG systems, the \textit{core pipeline fault} is the most widespread exhibited issue, consisting of 64\%, since the related options, if contain issues, often cause severe results such as crashes; while for the latter, \textit{quality fault} is more commonly observed, as it is often more critical than performance. 

There are also discrepancies among the systems, e.g., \textsc{Quivr}---external tools-heavy RAG systems---show particularly high proportions of core pipeline faults, as their orchestration logic is strongly coupled with configuration. \textsc{FastGPT}, a lightweight RAG system, exhibits more behavior faults.

\begin{tcolorbox}[enhanced,
left=0pt,
right=0pt,
top=2pt,
bottom=0pt,
boxrule=0.2mm, title=\textbf{\textit{Finding 1}},
colframe=black,colback=white,colbacktitle=white,
coltitle=black,attach boxed title to top center=
{yshift=-0.25mm-\tcboxedtitleheight/2,yshifttext=2mm-\tcboxedtitleheight/2},
boxed title style={boxrule=-0.2mm,
frame code={ \path[tcb fill frame] ([xshift=-3mm]frame.west)
-- (frame.north west) -- (frame.north east) -- ([xshift=3mm]frame.east)
-- (frame.south east) -- (frame.south west) -- cycle; },
interior code={ \path[tcb fill interior] ([xshift=-2.1mm]interior.west)
-- (interior.north west) -- (interior.north east)
-- ([xshift=2.1mm]interior.east) -- (interior.south east) -- (interior.south west)
-- cycle;} }]
Configuration issues in RAG systems are more predominantly manifested as {functional execution faults} (86\%/88\% ) rather than {behavior faults} (14\%/12\%). Faults related to the core pipeline and the answers' quality are the most common. 


\end{tcolorbox}


\subsubsection{Symptoms-Stage Relations}


\autoref{tab:symptom_stage} reveals two critical observations on the symptoms-stage relations: (1) the stage \textit{initialization} accounts for 24\%/22\% of the issues, meaning that users frequently fail to even start the system. (2) The issues are more associated with either the early or late stage. That is, we see that stage \textit{indexing} and \textit{generation} own 32\%/30\% and 27\%/32\% of the issues, respectively, which are considerably higher than any other intermediate stages.

\begin{table}[t!]
\centering
\caption{Overall symptoms proportion over pipeline stages. The highest is highlighted in \colorbox[HTML]{E6F0FF}{blue}.}
\label{tab:symptom_stage}

\begin{adjustbox}{max width=0.70\linewidth,center}
\begin{tabular}{llll}
\toprule
\textbf{Pipeline Stage} & \textbf{Count} & \textbf{Weighted} & \textbf{Unweighted} \\
\midrule

\rowcolors{1}{white}{gray!15}

S0: Initialization      & 153          & 24\%   & 22\% \\
S1: Indexing          & \cellcolor{highlight}208 & \cellcolor{highlight}32\%  & 30\% \\
S2: Query Transformation      & 23           & 3\%  & 3\% \\
S3: Retrieval                 & 70           & 11\% & 11\%\\
S4: Post-Retrieval Processing & 23           & 3\%  & 2\%\\
S5: Generation        & 177 & 27\% & \cellcolor{highlight}32\%\\

\bottomrule
\end{tabular}
\end{adjustbox}
\end{table}




\begin{tcolorbox}[enhanced,boxrule=0.2mm, 
left=0pt,
right=0pt,
top=2pt,
bottom=0pt,title=\textbf{\textit{Finding 2}},
colframe=black,colback=white,colbacktitle=white,
coltitle=black,attach boxed title to top center=
{yshift=-0.25mm-\tcboxedtitleheight/2,yshifttext=2mm-\tcboxedtitleheight/2},
boxed title style={boxrule=-0.2mm,
frame code={ \path[tcb fill frame] ([xshift=-3mm]frame.west)
-- (frame.north west) -- (frame.north east) -- ([xshift=3mm]frame.east)
-- (frame.south east) -- (frame.south west) -- cycle; },
interior code={ \path[tcb fill interior] ([xshift=-2.1mm]interior.west)
-- (interior.north west) -- (interior.north east)
-- ([xshift=2.1mm]interior.east) -- (interior.south east) -- (interior.south west)
-- cycle;} }]
Configuration issues in RAG systems predominantly surface at three critical stages: 
the \textit{initialization} (24\%/22\%), \textit{indexing} (32\%/30\%) and \textit{generation} (27\%/32\%).

\end{tcolorbox}

\begin{figure}[t!]
    \centering
    \includegraphics[width=0.90\linewidth]{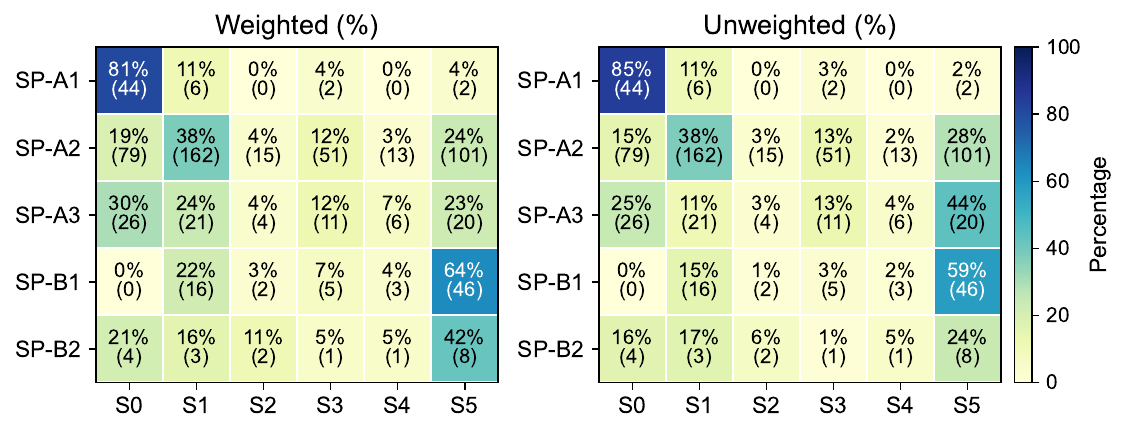}
    \caption{Detailed symptoms \% per category over stages.}
    \label{fig:rq1_heatmap}
\end{figure}

To take a closer look, as shown in \autoref{fig:rq1_heatmap}, there is a distinct temporal separation in symptom manifestation. 54\%-92\%/36\%-96\% of the \textit{functional execution fault} occurs at the first two stages, often blocking further execution. In contrast, the \textit{behavior fault} is related to the late stages, with 42\%-64\%/24\%-59\% at the \textit{generation} stage, manifesting after the full pipeline has executed. This is because the early stages can easily exhibit errors related to preparations, while the late stage is often where the problematic answers are observed.




\begin{tcolorbox}[enhanced,boxrule=0.2mm, 
left=0pt,
right=0pt,
top=2pt,
bottom=0pt,title=\textbf{\textit{Finding 3}},
colframe=black,colback=white,colbacktitle=white,
coltitle=black,attach boxed title to top center=
{yshift=-0.25mm-\tcboxedtitleheight/2,yshifttext=2mm-\tcboxedtitleheight/2},
boxed title style={boxrule=-0.2mm,
frame code={ \path[tcb fill frame] ([xshift=-3mm]frame.west)
-- (frame.north west) -- (frame.north east) -- ([xshift=3mm]frame.east)
-- (frame.south east) -- (frame.south west) -- cycle; },
interior code={ \path[tcb fill interior] ([xshift=-2.1mm]interior.west)
-- (interior.north west) -- (interior.north east)
-- ([xshift=2.1mm]interior.east) -- (interior.south east) -- (interior.south west)
-- cycle;} }]
\textit{Functional execution fault} often occurs much earlier than the \textit{behavior fault} at the pipeline stages of RAG systems. 

\end{tcolorbox}

\subsection{RQ2: Root Causes of Configuration Issues}
\label{sec:rq2}

\begin{table}[t!]
\centering
\caption{Root cause proportion by systems/categories. The same format as Table~\ref{tab:symptom_distribution}.}
\label{tab:root_cause_distribution}

\setlength{\tabcolsep}{0.8mm}
\begin{adjustbox}{max width=0.92\linewidth,center}
\begin{tabular}{llllll|llll|r}
\toprule
\textbf{System} & \textbf{RC-A1} & \textbf{RC-A2} & \textbf{RC-A3} & \textbf{RC-A4} & \textbf{RC-A5} & \textbf{RC-B1} & \textbf{RC-B2} & \textbf{RC-B3} & \textbf{RC-B4} & \textbf{Total} \\
\midrule

\rowcolors{1}{white}{gray!15}

\textsc{RAGFlow}          & 18\% (38) & \cellcolor{highlight}28\% (61) & 10\% (21) & 26\% (56) & 3\% (7)   & 5\% (10)  & 1\% (3)  & 7\% (16) & 2\% (5) & 217 \\
\textsc{DB-GPT}           & 12\% (14) & 13\% (16) & 13\% (16) & \cellcolor{highlight}32\% (38) & 12\% (14) & 8\% (10)  & 2\% (2)  & 8\% (9)  & 1\% (1) & 120 \\
\textsc{FastGPT}          & 5\% (4)   & 8\% (6)   & 0\% (0)   & 28\% (21) & 3\% (2)   & \cellcolor{highlight}38\% (28) & 4\% (3)  & 14\% (10) & 0\% (0)     & 74  \\
\textsc{GraphRAG}         & \cellcolor{highlight}33\% (21) & 19\% (12) & 9\% (6)   & 19\% (12) & 2\% (1)   & 6\% (4)   & 3\% (2)  & 9\% (6)  & 0\% (0)     & 64  \\
\textsc{LightRAG}         & 6\% (4)   & 17\% (11) & 6\% (4)   & \cellcolor{highlight}40\% (25) & 0\% (0)        & 14\% (9)  & 0\% (0)       & 16\% (10) & 0\% (0)     & 63  \\
\textsc{Obsidian-copilot} & 29\% (10) & 3\% (1)   & 0\% (0)   & 17\% (6)  & 6\% (2)   & \cellcolor{highlight}34\% (12) & 3\% (1)  & 9\% (3)  & 0\% (0)     & 35  \\
\textsc{Private-GPT}      & 6\% (2)   & 0\% (0)        & 10\% (3)  & 23\% (7)  & 10\% (3)  & \cellcolor{highlight}42\% (13) & 3\% (1)  & 6\% (2)  & 0\% (0)     & 31  \\
\textsc{Quivr}            & 9\% (2)   & 9\% (2)   & 0\% (0)   & \cellcolor{highlight}59\% (13) & 0\% (0)        & 18\% (4)  & 0\% (0)       & 5\% (1)  & 0\% (0)     & 22  \\
\textsc{DocsGPT}          & 24\% (5)  & 0\% (0)        & 0\% (0)        & 24\% (5)  & 10\% (2)  & \cellcolor{highlight}29\% (6)  & 5\% (1)  & 10\% (2) & 0\% (0)     & 21  \\
\textsc{HippoRAG}         & 14\% (1)  & 14\% (1)  & 0\% (0)        & 14\% (1)  & 0\% (0)        & \cellcolor{highlight}43\% (3)  & 0\% (0)       & 14\% (1) & 0\% (0)     & 7   \\

\midrule
\rowcolor{white}
\textbf{Weighted} & 15\% (101) & 17\% (110) & 8\% (50) & \cellcolor{highlight}28\% (184) & 5\% (31) & 15\% (99) & 2\% (13) & 9\% (60) & 1\% (6) & 654 \\
\textbf{Unweighted} & 16\% (101) & 11\% (110) & 5\% (50) & \cellcolor{highlight}28\% (184) & 4\% (31) & 24\% (99) & 2\% (13) & 10\% (60) & 0\% (6) & 654 \\
\bottomrule
\end{tabular}
\end{adjustbox}
\end{table}

\subsubsection{Commonality of Root Causes}

As shown in \autoref{tab:root_cause_distribution}, \textit{configuration bug} (RC-A1 to RC-A5) dominates the root causes, accounting for 73\%/64\% of the weighted/unweighted percentage, while \textit{user misconfiguration} accounts for the remaining 27\%/36\%. Notably, \textit{external system adaptation bug} (RC-A4) emerges as the most prevalent root cause of configuration bugs. This reflects the fundamental nature of RAG systems as {integration-centric architectures}: they act as glue code that orchestrates multiple heterogeneous components with evolving interfaces and implicit assumptions. As a result, configuration bugs frequently surface at those boundaries. 



As for \textit{user misconfiguration}, the \textit{external configuration mistake} (RC-B1) is the most common cause. This suggests that user mistakes are most likely to arise from configuring external artefacts that do not meet the RAG system’s needs.

There is also a notable system discrepancy: highly-integrated and cloud-native assistants, e.g., \textsc{Quivr}, predominantly suffer from external system adaptation bugs due to their extensive reliance on third-party APIs, whereas local-deployment or lightweight systems, e.g., \textsc{Private-GPT} and \textsc{FastGPT} exhibit a significantly higher proportion of external configuration mistakes (38\%–42\%) as users frequently struggle with local environment and model provisioning.



\begin{tcolorbox}[enhanced,boxrule=0.2mm, 
left=0pt,
right=0pt,
top=2pt,
bottom=0pt,title=\textbf{\textit{Finding 4}},
colframe=black,colback=white,colbacktitle=white,
coltitle=black,attach boxed title to top center=
{yshift=-0.25mm-\tcboxedtitleheight/2,yshifttext=2mm-\tcboxedtitleheight/2},
boxed title style={boxrule=-0.2mm,
frame code={ \path[tcb fill frame] ([xshift=-3mm]frame.west)
-- (frame.north west) -- (frame.north east) -- ([xshift=3mm]frame.east)
-- (frame.south east) -- (frame.south west) -- cycle; },
interior code={ \path[tcb fill interior] ([xshift=-2.1mm]interior.west)
-- (interior.north west) -- (interior.north east)
-- ([xshift=2.1mm]interior.east) -- (interior.south east) -- (interior.south west)
-- cycle;} }]
\textit{Configuration bug} (73\%/64\%) is the primary cause of configuration issues in RAG systems against \textit{user misconfiguration} (27\%/36\%). Notably, \textit{external system adaptation bug} is the most predominant configuration bug, while the most common user mistakes are from configuring external artefacts.

\end{tcolorbox}

\subsubsection{Root Cause-Stage Relation}


\begin{table}[t!]
\centering
\caption{Overall root cause proportion over pipeline stages. The highest is highlighted in \colorbox[HTML]{E6F0FF}{blue}.}
\label{tab:rootcause_stage}

\begin{adjustbox}{max width=0.70\linewidth,center}
\begin{tabular}{llll}
\toprule
\textbf{Pipeline Stage} & \textbf{Count} & \textbf{Weighted} & \textbf{Unweighted} \\
\midrule

\rowcolors{1}{white}{gray!15}

S0: Initialization                & \cellcolor{highlight}255 & \cellcolor{highlight}39\% & \cellcolor{highlight}40\%\\
S1: Indexing                      & 177 & 27\% & 25\%\\
S2: Query Transformation          & 27           & 4\%  & 3\%\\
S3: Retrieval                     & 43           & 7\%  & 5\% \\
S4: Post-Retrieval Processing     & 23           & 3\%  & 2\%\\
S5: Generation                    & 129          & 20\% & 24\% \\

\bottomrule
\end{tabular}
\end{adjustbox}
\end{table}

\begin{figure}[t!]
    \centering
    \includegraphics[width=0.90\linewidth]{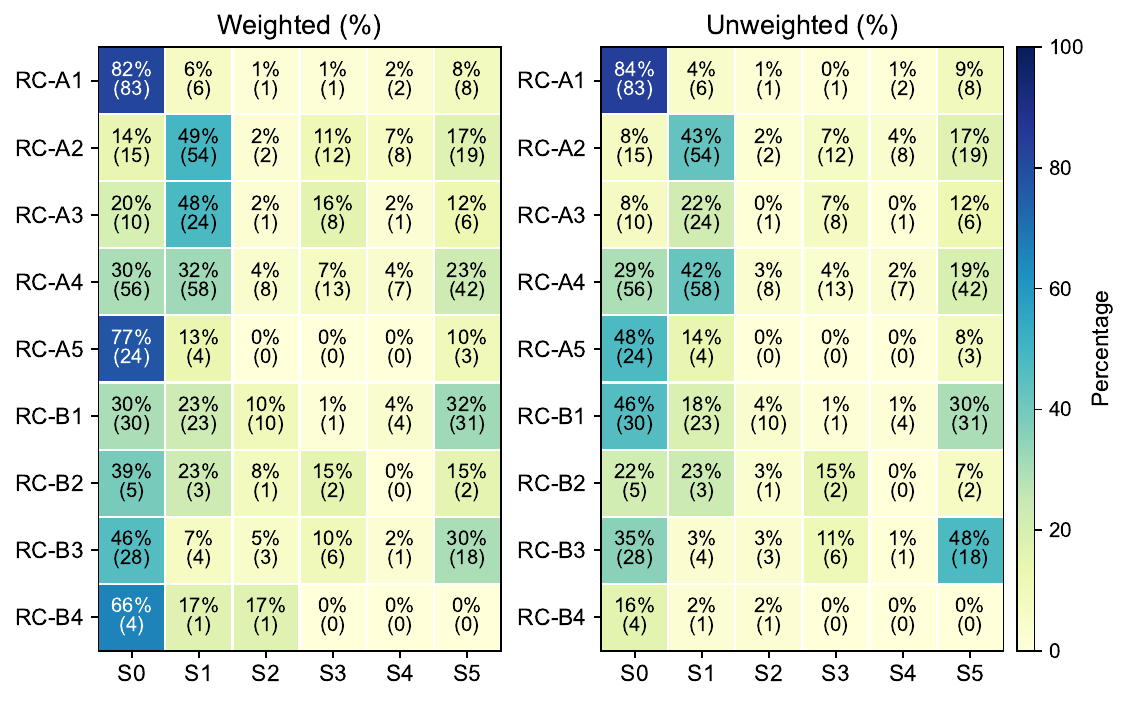}
    \caption{Detailed root cause \% per category over stages.}
    \label{fig:rq2_heatmap}
\end{figure}

We further analyze at which RAG stage the root causes are identified. As shown in \autoref{tab:rootcause_stage}, the majority of root causes lie in the early stage (\textit{initialization} and \textit{indexing}), which together account for over 66\%/65\% of the cases. The same can be observed over the sub-categories (\autoref{fig:rq2_heatmap}). This highlights that many configuration issues are rooted in system setup and data preparation. 




\begin{tcolorbox}[enhanced,boxrule=0.2mm, 
left=0pt,
right=0pt,
top=2pt,
bottom=0pt,title=\textbf{\textit{Finding 5}},
colframe=black,colback=white,colbacktitle=white,
coltitle=black,attach boxed title to top center=
{yshift=-0.25mm-\tcboxedtitleheight/2,yshifttext=2mm-\tcboxedtitleheight/2},
boxed title style={boxrule=-0.2mm,
frame code={ \path[tcb fill frame] ([xshift=-3mm]frame.west)
-- (frame.north west) -- (frame.north east) -- ([xshift=3mm]frame.east)
-- (frame.south east) -- (frame.south west) -- cycle; },
interior code={ \path[tcb fill interior] ([xshift=-2.1mm]interior.west)
-- (interior.north west) -- (interior.north east)
-- ([xshift=2.1mm]interior.east) -- (interior.south east) -- (interior.south west)
-- cycle;} }]
All types of root causes of configuration issues in RAG are often related to early RAG pipeline stages (66\%/65\% cases).

\end{tcolorbox}

\subsubsection{Configuration Options-Stage Relations}

\begin{table}[t!]
\centering
\caption{Proportion of $607$ error-prone configuration option over stages. The highest is highlighted in \colorbox[HTML]{E6F0FF}{blue}.}
\label{tab:parameter_distribution}

\begin{adjustbox}{max width=0.70\linewidth,center}
\begin{tabular}{llll}
\toprule
\textbf{Pipeline Stage} & \textbf{Count} & \textbf{Weighted} & \textbf{Unweighted} \\
\midrule

\rowcolors{1}{white}{gray!15}


S0: Initialization & 131 & 21\% & 22\%\\
S1: Indexing & \cellcolor{highlight}239 & \cellcolor{highlight}39\% & \cellcolor{highlight}38\% \\
S2: Query Transformation & 25 & 4\% & 3\%\\
S3: Retrieval & 40 & 7\% & 7\%\\
S4: Post-Retrieval Processing & 34 & 6\% & 4\% \\
S5: Generation & 138 & 23\% & 26\%\\

\bottomrule
\end{tabular}
\end{adjustbox}
\end{table}

To understand the granularity of configuration risks, we extracted the specific configuration options of root causes in the issues. Notblity, to prevent repeatedly reported problems from disproportionately increasing the frequency of a configuration option, we deduplicated the option-level data within each repository. Specifically, when multiple issues from the same repository involved the same configuration option, that option was counted only once. The same option name appearing in different repositories was retained separately because its implementation, semantics, and affected component may differ across systems. Consequently, the resulting 607 repository--option pairs characterize the breadth of distinct issue-linked configuration options rather than the raw frequency of issue reports. 

In total, we identify {607 unique options} across the 654 issues (an issue may involve multiple options). As in \autoref{tab:parameter_distribution}, we say that: 


\begin{tcolorbox}[enhanced,boxrule=0.2mm,
left=0pt,
right=0pt,
top=2pt,
bottom=0pt,title=\textbf{\textit{Finding 6}},
colframe=black,colback=white,colbacktitle=white,
coltitle=black,attach boxed title to top center=
{yshift=-0.25mm-\tcboxedtitleheight/2,yshifttext=2mm-\tcboxedtitleheight/2},
boxed title style={boxrule=-0.2mm,
frame code={ \path[tcb fill frame] ([xshift=-3mm]frame.west)
-- (frame.north west) -- (frame.north east) -- ([xshift=3mm]frame.east)
-- (frame.south east) -- (frame.south west) -- cycle; },
interior code={ \path[tcb fill interior] ([xshift=-2.1mm]interior.west)
-- (interior.north west) -- (interior.north east)
-- ([xshift=2.1mm]interior.east) -- (interior.south east) -- (interior.south west)
-- cycle;} }]
The complexity of RAG configuration is unevenly distributed: the \textit{indexing} stage involves the highest volume of error-prone options ({39\%/38\%}), such as embedding settings. 

\end{tcolorbox}

\subsection{RQ3: Stage Gap and Resolving Costs}
\label{sec:rq3}

\textbf{RQ3} investigates how costly it is to resolve the configuration issues of RAG systems in practice, depending on the deviation between the stage of symptoms and root causes. To that end, we compute the stage gap $\mathcal{G}$ of each issue:
\begin{equation}
    \mathcal{G} = |\mathcal{S} - \mathcal{R}|
\end{equation}
whereby $\mathcal{S}$ and $\mathcal{R}$ denote the stage level for symptom and root cause, respectively. Clearly, a $\mathcal{G} = 0$ indicates the issue manifests at the same stage where it was introduced. A $\mathcal{G} > 0$ indicates cross-stage propagation, where the configuration issues are propagated through other stages before revealing a symptom. Note that the cause might be related to a stage earlier or later than the stage of the symptom.






For example, the issue $\#144$ for \textsc{Obsidian-copilot} is a misconfiguration issue. Here, the user configured the RAG system to use the \textit{``Active Note''} context mode, mistakenly believing it would feed the entire document as full context. However, the system's underlying logic uses similarity-based retrieval. As such, the stage of root cause is \textit{retrieval}, as the user's misunderstanding led to selecting a retrieval strategy unsuitable for the task. Yet, the symptom only became observable at the \textit{generation} stage, when the LLM produced a low-quality response. This results in a $\mathcal{G} = |5 - 3| = 2$. 

\subsubsection{Prevalence and Patterns of Stage Gap}



\begin{table}[t!]
\centering
\caption{Stage gap proportion across root cause categories; W and U denote weighted and unweighted percentage, respectively. The highest is highlighted in \colorbox[HTML]{E6F0FF}{blue}.}
\label{tab:gap_distribution}

\begin{adjustbox}{max width=0.85\linewidth,center}
\begin{tabular}{llll}
\toprule
\textbf{Stage Gap} & 
\makecell[c]{\textbf{Overall} \\ \small \textbf{[Count (W/U\%)]}} & 
\makecell[c]{\textbf{RC-A} \\ \small \textbf{[Count (W/U\%)]}} & 
\makecell[c]{\textbf{RC-B} \\ \small \textbf{[Count (W/U\%)]}} \\
\midrule

Gap = 0   & \cellcolor{highlight}440 (67\%/64\%) &\cellcolor{highlight}332 (70\%/67\%) & \cellcolor{highlight}108 (61\%/57\%) \\
Gap = 1           & 93 (15\%/13\%)  & 61 (13\%/8\%)   & 32 (18\%/21\%)  \\
Gap = 2           & 27 (4\%/5\%)    & 19 (4\%/3\%)    & 8 (5\%/5\%)     \\
Gap = 3           & 19 (3\%/5\%)    & 17 (3\%/9\%)    & 2 (1\%/2\%)     \\
Gap = 4           & 15 (2\%/2\%)    & 9 (2\%/2\%)     & 6 (3\%/2\%)     \\
Gap = 5           & 60 (9\%/11\%)   & 38 (8\%/11\%)   & 22 (12\%/13\%)  \\

\midrule

Total Gap $> 0$ & 214 (33\%/36\%) & 144 (30\%/33\%) & 70 (39\%/43\%) \\
\bottomrule
\end{tabular}
\end{adjustbox}
\end{table}

\begin{figure}[t!]
    \centering
    \includegraphics[width=0.90\linewidth]{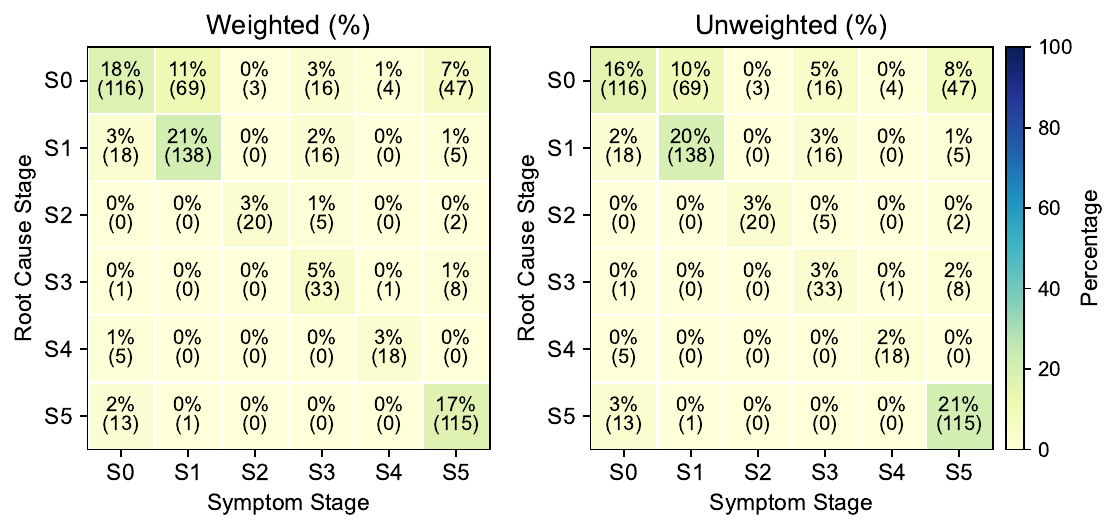}
    \caption{Stage matrix for symptom and root cause.}
    \label{fig:rq3_heatmap}
\end{figure}


As summarized in \autoref{tab:gap_distribution}, 33\%/36\% (weighted/unweighted) of the configuration issues exhibit a non-zero stage gap between root cause and symptom. This indicates that nearly one-third of configuration issues in RAG systems cannot be diagnosed by inspecting the stage where the failure is observed alone. Notably, this gap is present in both primary categories of the root cause, suggesting that cross-stage reasoning is a challenge in debugging the configuration issues for RAG systems.

\autoref{fig:rq3_heatmap} reveals that for those configuration issues with deviated stages, most commonly the symptoms can be at either \textit{indexing} (S1) or \textit{generation} (S5) while the root cause is associated with \textit{initialization} (S0) stage---the configuration issues introduced during system \textit{initialization} only become apparent as abnormalities propagate to the \textit{indexing} and \textit{generation}.




\begin{tcolorbox}[enhanced,boxrule=0.2mm, 
left=0pt,
right=0pt,
top=2pt,
bottom=0pt,title=\textbf{\textit{Finding 7}},
colframe=black,colback=white,colbacktitle=white,
coltitle=black,attach boxed title to top center=
{yshift=-0.25mm-\tcboxedtitleheight/2,yshifttext=2mm-\tcboxedtitleheight/2},
boxed title style={boxrule=-0.2mm,
frame code={ \path[tcb fill frame] ([xshift=-3mm]frame.west)
-- (frame.north west) -- (frame.north east) -- ([xshift=3mm]frame.east)
-- (frame.south east) -- (frame.south west) -- cycle; },
interior code={ \path[tcb fill interior] ([xshift=-2.1mm]interior.west)
-- (interior.north west) -- (interior.north east)
-- ([xshift=2.1mm]interior.east) -- (interior.south east) -- (interior.south west)
-- cycle;} }] 
Nearly one-third of RAG configuration issues exhibit a mismatch between the stages of symptom and root cause. This often happens when root problems at the \textit{initialization} cause the symptoms at either \textit{indexing} or \textit{generation} stage.

\end{tcolorbox}

\subsubsection{Impacts on Resolving Costs}

To understand the impacts of the stage gap, we quantify the duration required to resolve each configuration issue. Specifically, for each issue, the start time is defined as the issue creation timestamp (\texttt{created\_at}). 
The end time is determined depending on two cases: 

\begin{itemize}
    \item if the issue is relevant to a fixing pull request, we use the pull request merge time (\texttt{merged\_at}); 
    \item otherwise, we use the issue closing time (\texttt{closed\_at}). 
\end{itemize}


The resolving cost is calculated as the elapsed time between the start and end timestamps. From \autoref{tab:gap_cost_horizontal}, we observe that there seems to be a complex relationship between the stage gap and the resolving cost. We hence take two further analyses.



\begin{table}[t!]
\centering
\caption{Median resolving cost over stage gap.}
\label{tab:gap_cost_horizontal}

\begin{adjustbox}{width=0.6\columnwidth,center}
\begin{tabular}{l | cccccc}
\toprule
\textbf{Stage gap } & 0 & 1 & 2 & 3 & 4 & 5 \\
\midrule

\textbf{Median cost (h)} & 66.0 & 35.1 & 102.4 & 357.2 & 47.1 & 167.2 \\

\bottomrule
\end{tabular}
\end{adjustbox}
\end{table}

\textbf{Correlation Analysis.} Our initial hypothesis is that there could be a monotonic relationship between those two. To confirm this, we apply Spearman correlation $r$ on the stage gap ($\mathcal{G}\in[0,5]$) to the resolving cost for all configuration issues. The result indicates that they have a trivial correlation ($r \approx 0.068$, $p$-value $\approx 0.08$). As such, a larger stage gap might not mean that the configuration issue takes longer to resolve. 



\textbf{Grouping Analysis.} Based on the correlation results, we revised our hypothesis: there might exist a \textit{``complexity threshold''} where cross-stage debugging of configuration issues becomes significantly harder. To that end, we apply the {Mann-Whitney U test} to compare the resolving costs between two groups of issues separated by different thresholds, and we found that only when $\mathcal{G}=2$ yields statistically significant differences between the groups in the resolving cost: we evaluated each feasible cutoff $c\in\{1,2,3,4,5\}$ by comparing issues with $\mathcal{G}<c$ against those with $\mathcal{G}\geq c$. The cutoffs $c=1$, $c=4$, and $c=5$ did not produce statistically significant differences in resolving costs, whereas both $c=2$ and $c=3$ produced significant differences. We selected $c=2$ for the reported grouping because, compared with $c=3$, it yielded a smaller overall $p$-value ($0.0025$ versus $0.0059$), a larger cost multiplier for configuration bugs ($2.83\times$ versus $2.77\times$), and a more balanced high-gap sample ($N=121$ versus $N=94$). Therefore, we dichotomize the configuration issues for RAG systems as: 
\begin{itemize}
    \item \textbf{Low gap ($\mathcal{G}\in[0,1]$)}: Issues that are localized or manifest in the same or immediate adjacent stage.
    \item \textbf{High gap ($\mathcal{G}\in[2,5]$)}: Issues that propagate more stages.
\end{itemize}


\noindent As shown in \autoref{tab:gap_MWU}, we observe:

\begin{itemize}
    \item \textbf{Overall impact}: For all configuration issues, those with the high gap have a median resolving cost of {160.6 hours}, which is {2.48$\times$} higher than that of those within a low gap (64.7 hours) with statistical significance ($p < 0.01$).
    
    \item \textbf{Configuration bugs (RC-A) as root causes}: A deeper breakdown reveals that the large resolving cost difference is primarily driven by configuration bugs: the cost multiplier rises to {2.83$\times$} ($p < 0.01$) in the high gap issues. This implies that when internal code logic fails and propagates downstream, developers face substantial difficulty in tracing the RAG pipeline to the root causes.
    
    \item \textbf{Misconfiguration (RC-B) as root causes}: For user misconfigurations, although the median cost increases (1.87$\times$) in the high gap group, the difference is not statistically significant. This confirms that while user misconfiguration issues exhibit large stage gaps, they are ``cheaper'' to resolve (often a single option update).
\end{itemize}






\begin{table}[t!]
\centering
\caption{Statistical difference of resolving costs (hours) on RAG configuration issues with low and high gap.}
\label{tab:gap_MWU}

\setlength{\tabcolsep}{2.5pt}
\begin{adjustbox}{max width=0.72\linewidth,center}
\begin{tabular}{llllllr}
\toprule
\multirow{2}{*}{\textbf{Group}} & \multicolumn{2}{c}{\textbf{Low Gap}} & \multicolumn{2}{c}{\textbf{High Gap}} & \multicolumn{2}{c}{\textbf{U Test}} \\
\cmidrule(lr){2-3} \cmidrule(lr){4-5} \cmidrule(lr){6-7}
 & \textbf{Count} & \textbf{Median} & \textbf{Count} & \textbf{Median} & \textbf{Increase} & \textbf{$p$-value} \\
\midrule

\rowcolors{1}{white}{gray!15}

\textbf{Overall}      & 533 & 64.7 & 121 & 160.6 & 2.48$\times$ & 0.003 \\
\textbf{RC-A}: bugs  & 393 & 58.5 & 83  & 165.2 & 2.83$\times$ & 0.002 \\
\textbf{RC-B}: misconfig.  & 140 & 78.3 & 38  & 146.1 & 1.87$\times$          & 0.273 \\

\bottomrule
\end{tabular}
\end{adjustbox}
\end{table}


Therefore, we conclude:

\begin{tcolorbox}[enhanced,boxrule=0.2mm, 
left=0pt,
right=0pt,
top=2pt,
bottom=0pt,title=\textbf{\textit{Finding 8}},
colframe=black,colback=white,colbacktitle=white,
coltitle=black,attach boxed title to top center=
{yshift=-0.25mm-\tcboxedtitleheight/2,yshifttext=2mm-\tcboxedtitleheight/2},
boxed title style={boxrule=-0.2mm,
frame code={ \path[tcb fill frame] ([xshift=-3mm]frame.west)
-- (frame.north west) -- (frame.north east) -- ([xshift=3mm]frame.east)
-- (frame.south east) -- (frame.south west) -- cycle; },
interior code={ \path[tcb fill interior] ([xshift=-2.1mm]interior.west)
-- (interior.north west) -- (interior.north east)
-- ([xshift=2.1mm]interior.east) -- (interior.south east) -- (interior.south west)
-- cycle;} }]
There is no monotonic correlation between stage gap size and resolving cost. Instead, we observe a \textit{``threshold effect''}: costs spike significantly when issues have $\mathcal{G}\ge 2$, leading to largely deviated costs mainly due to configuration bugs. 


\end{tcolorbox}
\subsection{RQ4: Fix Pattern}
\label{sec:rq4}

We now focus on studying the 345 configuration issues for RAG systems, which are linked with pull requests (including both configuration bugs and user misconfiguration). As shown in \autoref{tab:fix_patterns}, we see that for both weighted and unweighted percentages, the \textit{component logic and algorithm}
(FP2, 40\%/43\%) and \textit{data interaction and mapping} (FP1, 28\%/26\%) are among the most common fix patterns. However, they are not necessarily those that require the greatest effort in terms of fixing. Instead, patterns that involve deeper system-level intervention, such as  \textit{lifecycle and resource scheduling} (FP6) exhibit noticeably higher median fix durations.

Specifically, graph-based and multi-source integrated systems, e.g., \textsc{GraphRAG} and \textsc{Quivr} require \textit{data interaction and mapping} fix to resolve complex data structure mismatches, whereas other systems heavily rely on \textit{component logic and algorithm fix} to repair their routing and execution flows.

\begin{figure}[t!]
    \centering
    \includegraphics[width=0.90\linewidth]{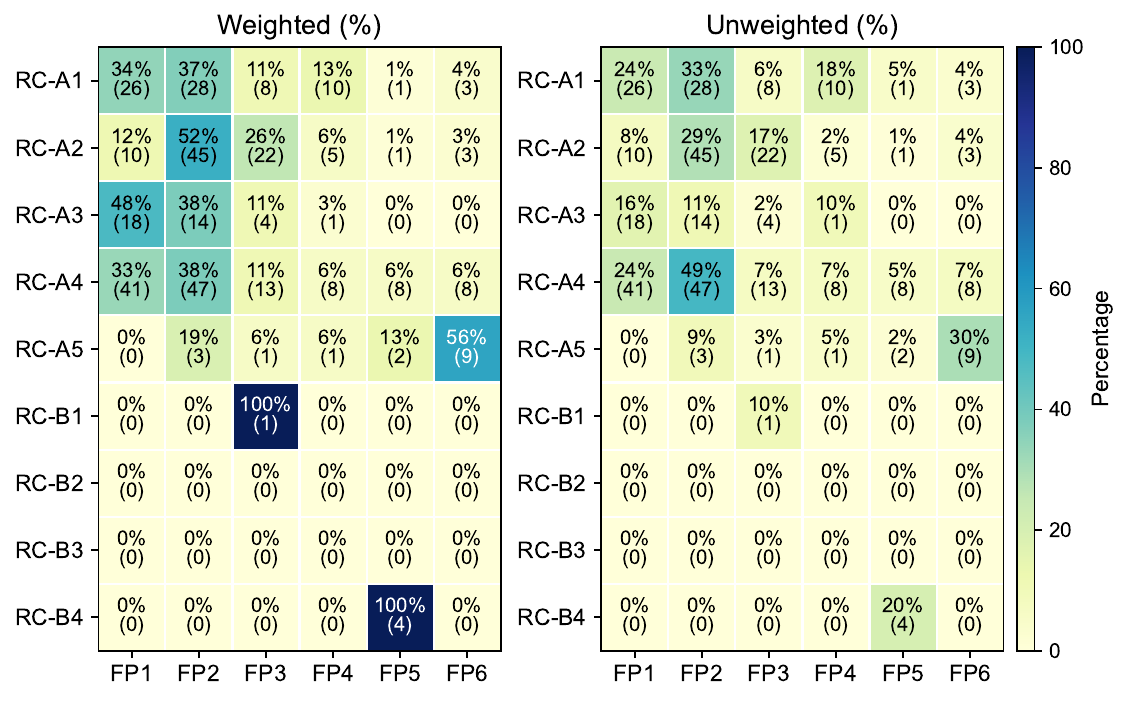}
    \caption{Distribution of fix patterns across different root causes.}
    \label{fig:rq4_heatmap}
\end{figure}

From \autoref{fig:rq4_heatmap}, we see several patterns between root causes and fix patterns, where the configuration bugs are often linked to \textit{data interaction and mapping} fixes (FP1) and \textit{component logic and algorithm fixes} (FP2). Interestingly, misconfiguration can also lead to fixes, mainly on the improved error checking/handling (FP3) and document/manual updates (FP5).

\begin{table}[t]
\centering
\caption{Fix pattern proportion by systems/categories. The same format as Table~\ref{tab:symptom_distribution}.}
\label{tab:fix_patterns}

\setlength{\tabcolsep}{1mm}
\begin{adjustbox}{max width=0.90\linewidth,center}
\begin{tabular}{lllllll|r}
\toprule
\textbf{System} & \textbf{FP1} & \textbf{FP2} & \textbf{FP3} & \textbf{FP4} & \textbf{FP5} & \textbf{FP6} & \textbf{Total} \\
\midrule
\rowcolors{1}{white}{gray!15}

\textsc{RAGFlow}          & 25\% (40) & \cellcolor{highlight}40\% (64) & 21\% (34) & 6\% (9)  & 5\% (8)  & 4\% (6)  & 161 \\
\textsc{DB-GPT}           & 26\% (20) & \cellcolor{highlight}39\% (30) & 8\% (6)   & 8\% (6)  & 7\% (5)  & 12\% (9) & 76  \\
\textsc{GraphRAG}         & \cellcolor{highlight}50\% (14) & 32\% (9) & 11\% (3) & 4\% (1) & 0\% (0) & 4\% (1) & 28 \\
\textsc{LightRAG}         & 32\% (7)  & \cellcolor{highlight}36\% (8)  & 5\% (1)   & 18\% (4) & 0\% (0)       & 9\% (2)  & 22  \\
\textsc{FastGPT}          & 26\% (5)  & \cellcolor{highlight}47\% (9)  & 5\% (1)   & 11\% (2) & 0\% (0)       & 11\% (2) & 19  \\
\textsc{Obsidian-copilot} & 7\% (1)   & \cellcolor{highlight}60\% (9)  & 13\% (2)  & 20\% (3) & 0\% (0)       & 0\% (0)      & 15  \\
\textsc{DocsGPT}          & \cellcolor{highlight}40\% (4) & 20\% (2) & 10\% (1)  & 0\% (0)      & 20\% (2)  & 10\% (1) & 10  \\
\textsc{Quivr}            & \cellcolor{highlight}50\% (4) & 25\% (2) & 0\% (0)       & 0\% (0)      & 12\% (1)  & 12\% (1) & 8   \\
\textsc{HippoRAG}         & 0\% (0)       & \cellcolor{highlight}67\% (2)  & 33\% (1)  & 0\% (0)      & 0\% (0)       & 0\% (0)      & 3   \\
\textsc{Private-GPT}      & 0\% (0)       & \cellcolor{highlight}67\% (2)  & 0\% (0)       & 0\% (0)      & 0\% (0)       & 33\% (1) & 3   \\

\midrule
\rowcolor{white}
\textbf{Weighted}   & 28\% (95) & \cellcolor{highlight}40\% (137) & 14\% (49) & 7\% (25) & 5\% (16) & 6\% (23) & 345 \\
\textbf{Unweighted} & 26\% (95) & \cellcolor{highlight}43\% (137) & 11\% (49) & 7\% (25) & 4\% (16) & 9\% (23) & 345 \\
\textbf{Median cost (h)}    & 31.7     & 58.6      & 41.0     & 30.4    & 63.1    & \cellcolor{highlight}103.2   & \textbf{--} \\
\bottomrule
\end{tabular}
\end{adjustbox}
\end{table}

\begin{tcolorbox}[enhanced,boxrule=0.2mm,
left=0pt,
right=0pt,
top=2pt,
bottom=0pt,title=\textbf{\textit{Finding 9}},
colframe=black,colback=white,colbacktitle=white,
coltitle=black,attach boxed title to top center=
{yshift=-0.25mm-\tcboxedtitleheight/2,yshifttext=2mm-\tcboxedtitleheight/2},
boxed title style={boxrule=-0.2mm,
frame code={ \path[tcb fill frame] ([xshift=-3mm]frame.west)
-- (frame.north west) -- (frame.north east) -- ([xshift=3mm]frame.east)
-- (frame.south east) -- (frame.south west) -- cycle; },
interior code={ \path[tcb fill interior] ([xshift=-2.1mm]interior.west)
-- (interior.north west) -- (interior.north east)
-- ([xshift=2.1mm]interior.east) -- (interior.south east) -- (interior.south west)
-- cycle;} }]
The \textit{component logic and algorithm} (40\%/43\%) and \textit{data interaction and mapping} (28\%/26\%) are the most common patterns. Yet, those patterns that require deeper system-level interaction often require a higher cost to fix.

\end{tcolorbox}

\begin{tcolorbox}[enhanced,boxrule=0.2mm, 
left=0pt,
right=0pt,
top=2pt,
bottom=0pt,title=\textbf{\textit{Finding 10}},
colframe=black,colback=white,colbacktitle=white,
coltitle=black,attach boxed title to top center=
{yshift=-0.25mm-\tcboxedtitleheight/2,yshifttext=2mm-\tcboxedtitleheight/2},
boxed title style={boxrule=-0.2mm,
frame code={ \path[tcb fill frame] ([xshift=-3mm]frame.west)
-- (frame.north west) -- (frame.north east) -- ([xshift=3mm]frame.east)
-- (frame.south east) -- (frame.south west) -- cycle; },
interior code={ \path[tcb fill interior] ([xshift=-2.1mm]interior.west)
-- (interior.north west) -- (interior.north east)
-- ([xshift=2.1mm]interior.east) -- (interior.south east) -- (interior.south west)
-- cycle;} }]
Different root causes indeed exhibit distinct repair tendencies: configuration bugs are often related to data mapping and core logic fixes, while the misconfiguration can lead to improved error checking and document updates.


\end{tcolorbox}


\section{Insights for Configuration of RAG Systems}
\label{sec:insights}

\begingroup
\definecolor{boxblue}{HTML}{E6F0FF}
\definecolor{frameblue}{HTML}{315D9B}
\newtcolorbox{insightbox}[1]{
  colback=boxblue!35,
  colframe=frameblue,
  boxrule=0.8pt,
  arc=2pt,
  left=7pt,
  right=7pt,
  top=7pt,
  bottom=7pt,
  title=\textbf{#1}
}

Here, we discuss the insights derived from our empirical observations:

\subsection{For RAG System Developers}
\begin{insightbox}{Insight 1: Prioritize detection on stages other than query or retrieval-related ones.}

Most configuration issues in our dataset manifest as functional execution faults rather than only as degraded answer quality. Core pipeline faults are the most common symptom category, and symptoms concentrate at initialization, indexing, and generation. In contrast, query transformation, retrieval, and post-retrieval processing account for a smaller share of directly observable symptoms.

\end{insightbox}

\textbf{Guidance.}
Developers should treat configuration validation as a pipeline-level reliability mechanism rather than as a retrieval-only quality check. A practical validation suite should include at least three layers:

\begin{itemize}
    \item \textbf{Startup validation:} check whether required environment variables, model credentials, service endpoints, and database connections are available before the RAG service accepts user requests.
    \item \textbf{Indexing validation:} run a small indexing job with a known document and verify that chunking, embedding, metadata persistence, and vector-store insertion all complete successfully.
    \item \textbf{Generation validation:} run a short end-to-end query over a controlled fixture and verify that the configured prompt template, LLM endpoint, context window, and output parser are mutually compatible.
\end{itemize}

Retrieval metrics such as recall or ranking quality remain important, but they should not be the only oracle. Many RAG configuration failures prevent the system from reaching the retrieval-quality evaluation stage at all.

\textbf{Example.}
Suppose a RAG framework allows users to configure \texttt{embedding\_model}, \texttt{chunk\_size}, \texttt{vector\_store}, and \texttt{llm\_provider}. A developer can add a \texttt{rag doctor} command that performs the following checks before deployment: instantiate the embedding model, index one sample document, retrieve one chunk, and send a minimal prompt to the configured LLM. If the embedding dimension does not match the vector database collection, the command should fail during indexing with an explicit diagnostic, instead of letting the mismatch later appear as a retrieval or generation failure.


\begin{insightbox}{Insight 2: Focus debugging efforts on early-stage configurations.}

Most root causes originate from the initialization and indexing stages. This means that downstream failures are often caused by earlier configuration handling, resource binding, model loading, document processing, or index construction decisions.

\end{insightbox}

\textbf{Guidance.}
When a configuration issue is reported, developers should first reconstruct the early-stage configuration path:

\begin{enumerate}
    \item Identify which configuration source is authoritative, such as a YAML file, environment variable, command-line argument, UI setting, or database record.
    \item Confirm how the option is parsed, normalized, and propagated into the initialization or indexing component.
    \item Check whether the option silently falls back to a default value.
    \item Verify that derived artifacts, such as embedding dimensions, chunk metadata, index schema, and collection names, match the configured values.
\end{enumerate}

This debugging order is especially useful when the symptom appears later in the pipeline. A generation-stage error may still be caused by an incorrectly constructed index or by a missing initialization-time dependency.

\textbf{Example.}
A user reports that a generated answer ignores uploaded documents. Rather than starting from prompt tuning, the maintainer first checks whether the configured document parser and embedding model were used during indexing. The investigation reveals that the system accepted a custom embedding model in the UI, but the indexing worker still used the default model because the configuration was not propagated to the background job. The fix is to pass the selected model ID into the indexing job and record the model metadata with the generated index.

\begin{insightbox}{Insight 3: Cross-stage bug localization is necessary.}

Nearly one-third of configuration issues exhibit a non-zero stage gap between the root-cause stage and the symptom stage. Therefore, inspecting only the stage where the failure becomes visible is often insufficient.

\end{insightbox}

\textbf{Guidance.}
Developers should provide observability that connects configuration values to pipeline artifacts across stages. Useful mechanisms include:

\begin{itemize}
    \item a configuration provenance record that shows where each effective option came from;
    \item stage-level logs that include configuration fingerprints rather than only raw errors;
    \item metadata attached to indexed chunks, retrieved passages, and generated answers;
    \item an issue template that asks reporters for both runtime symptoms and upstream setup and indexing details.
\end{itemize}

The goal is to make it possible to trace how a configuration value affects later-stage behavior. Without this trace, maintainers may overfit the fix to the visible symptom and leave the upstream cause unchanged.

\textbf{Example.}
In a note-taking RAG application, a user chooses an ``active note'' mode expecting the whole note to be passed to the LLM. The system actually interprets this mode as a retrieval strategy over the note, and the symptom appears only when the LLM answers that it lacks enough context. The visible failure is in generation, but the root cause is a retrieval-stage configuration misunderstanding. A cross-stage trace should show: selected context mode $\rightarrow$ retrieval strategy $\rightarrow$ retrieved passages $\rightarrow$ final prompt. This trace would guide both documentation and UI improvements.

\begin{insightbox}{Insight 4: Prioritize the fix of issues with large stage gaps.}

Issues whose root causes and symptoms are separated by multiple pipeline stages incur substantially higher resolving costs. In our analysis, high-gap issues ($\mathcal{G}\ge2$) have a higher median resolving cost than low-gap issues, and the effect is particularly strong for system implementation defects.

\end{insightbox}

\textbf{Guidance.}
Maintainers should add stage-gap information to issue triage. Severity should still matter, but a non-crashing issue with a large stage gap may consume more debugging time than a localized crash with an obvious cause. A practical triage workflow can classify each report as:

\begin{itemize}
    \item \textbf{localized:} symptom and suspected root cause are in the same or adjacent stages;
    \item \textbf{propagated:} symptom appears two or more stages away from the suspected root cause;
    \item \textbf{unknown:} insufficient evidence, requiring reproduction logs or configuration provenance.
\end{itemize}

Propagated issues should receive earlier maintainer attention because they often require understanding multiple components and data contracts.

\textbf{Example.}
A generation-stage formatting failure may appear minor because the system still returns an answer. However, if the actual cause is that indexing stored metadata under the wrong key, then the issue spans indexing, retrieval, and generation. The maintainer should prioritize it over a localized prompt typo because the metadata mismatch can affect multiple retrieval strategies and future indexed data. A robust fix should migrate or normalize metadata at the indexing boundary, not merely patch the final prompt.

\begin{insightbox}{Insight 5: Specific tools are required for configuration repair.}

The most common fixes involve component logic and algorithms, followed by data interaction and mapping. Therefore, RAG configuration repair requires tools that understand component interactions, external dependencies, and pipeline data contracts, rather than only tools that edit configuration files.

\end{insightbox}

\textbf{Guidance.}
Repair support should be configuration-aware and component-aware. Useful tool features include:

\begin{itemize}
    \item checking whether configuration keys are consistently mapped across UI, API, worker, and storage layers;
    \item detecting incompatible component combinations, such as embedding models and existing vector-store schemas;
    \item generating regression tests from real configuration issues;
    \item suggesting code-level repairs when the system mishandles valid configurations.
\end{itemize}

This is different from a generic linter. Many observed repairs modify internal logic, interface mapping, or lifecycle management, which means the correct repair often belongs in framework code rather than in the user's configuration file.

\textbf{Example.}
A vector database adapter changes its expected field name from \texttt{metadata} to \texttt{payload}. Existing configurations still point to the same database, but retrieval fails because the RAG system maps chunk metadata to the old field. A configuration repair assistant should detect the adapter version, compare the expected schema with the configured mapping, and recommend an interface-level fix. The repair may update the adapter code and add a compatibility test for both field names.

\subsection{For RAG Systems Users}

\begin{insightbox}{Insight 6: Validate external artefacts before deployment.}

External configuration mistakes are the most common type of user misconfiguration. These mistakes involve external artefacts such as LLMs, embedding models, vector databases, re-rankers, local model files, credentials, and service endpoints.

\end{insightbox}

\textbf{Guidance.}
Users should validate external artefacts before relying on a full RAG deployment. A lightweight pre-deployment checklist should include:

\begin{itemize}
    \item the selected LLM endpoint is reachable and supports the configured context length, model name, authentication mode, and streaming option;
    \item the embedding model is available and its output dimension matches the vector database collection;
    \item the vector database is reachable, writable, and uses the expected metric, schema, and collection name;
    \item any re-ranker, parser, or external application integration uses a compatible API version.
\end{itemize}

The key is to test each external artefact independently before debugging the entire RAG pipeline.

\textbf{Example.}
A user switches from one embedding provider to another because the new model is cheaper. Before reusing the old vector collection, the user should check the new model's embedding dimension. If the old collection was built with 1536-dimensional vectors and the new model returns 1024-dimensional vectors, the correct action is to rebuild the index or create a separate collection. Otherwise, the system may fail during insertion or silently retrieve poor matches.

\begin{insightbox}{Insight 7: Pay close attention to configuration semantics and documentation.}

Some user misconfigurations arise from misunderstanding option semantics, constraints, deprecated parameters, or misleading documentation. A configuration can be syntactically valid but still conflict with the intended RAG behavior.

\end{insightbox}

\textbf{Guidance.}
Users should not treat defaults and option names as complete specifications. Before production deployment, they should inspect the semantic effects of critical options, especially those controlling context construction, chunking, retrieval mode, model selection, prompt templates, and resource limits. When documentation is ambiguous, users should run a small controlled example and inspect intermediate artifacts such as chunks, retrieved passages, and final prompts.

\textbf{Example.}
A configuration option named \texttt{context\_mode=active} may sound as if the system passes the entire active document to the LLM. In a particular implementation, it may instead mean that the active document is used as a search scope for similarity-based retrieval. A user preparing a legal or academic assistant should verify the final prompt for a test query. If only a few retrieved snippets appear, the user may need a different mode, a larger \texttt{top\_k}, or a custom prompt assembly strategy.

\subsection{For RAG Researchers}
\begin{insightbox}{Insight 8: Explore compatibility-sensitive configuration assistants.}

External system adaptation bugs are the most prevalent system-side root cause. This suggests that RAG systems need support for handling evolving APIs, changing model capabilities, incompatible vector database schemas, and heterogeneous component assumptions.

\end{insightbox}

\textbf{Guidance.}
Researchers can design assistants that reason about compatibility as a first-class property. Promising directions include:

\begin{itemize}
    \item mining dependency versions, model cards, API specifications, and adapter code to infer compatibility constraints;
    \item generating warnings when a selected LLM, embedding model, vector database, or re-ranker violates known constraints;
    \item learning from historical issue-fix pairs to identify fragile configuration combinations;
    \item continuously checking compatibility as external services evolve.
\end{itemize}

Such assistants should explain why a configuration is risky, not merely report that it is invalid. Many failures are caused by implicit assumptions that are not encoded in a single configuration schema.

\textbf{Example.}
A research prototype could monitor a RAG project's dependency file, environment variables, and adapter implementations. When the vector database client is upgraded, the assistant checks whether the configured collection schema and metadata mapping still match the adapter's expected API. If not, it suggests a migration plan and generates a regression test that indexes, retrieves, and generates over a small fixture.

\begin{insightbox}{Insight 9: Investigate pipeline-aware localization and repair techniques.}

Cross-stage issue propagation and the high resolving cost of large-gap issues indicate that RAG debugging tools should model the pipeline structure explicitly. Stage-agnostic bug localization may miss upstream causes that only become visible after several transformations.

\end{insightbox}

\textbf{Guidance.}
Future localization and repair techniques should represent RAG execution as a chain of configuration-dependent transformations: initialization, indexing, query transformation, retrieval, post-retrieval processing, and generation. A pipeline-aware tool can then reason over both control flow and data flow:

\begin{itemize}
    \item which configuration value affects which component;
    \item which intermediate artifact was produced under which configuration;
    \item where the first divergence from an expected artifact occurs;
    \item whether the repair should update user documentation, configuration validation, adapter mapping, or core pipeline logic.
\end{itemize}

\textbf{Example.}
Consider an issue where the final answer cites irrelevant documents. A pipeline-aware localizer should compare the execution trace against a known-good fixture: initialization loads the expected model, indexing creates chunks with correct metadata, query transformation preserves user intent, retrieval returns the expected candidates, post-retrieval filtering keeps valid passages, and generation uses the intended prompt template. If the first divergence appears in post-retrieval filtering because a configured metadata key was renamed during indexing, the repair should target metadata propagation rather than the LLM prompt.

\endgroup

\section{Threats to Validity}
\label{sec:threats}


To mitigate threats to \textbf{internal validity}, all authors are involved in the manual analysis for reaching agreements. We also calculate the Cohen’s $\kappa$ coefficient to evaluate inter-rater reliability iteratively---we stop only when there is $\kappa\geq0.7$ for all categories. Our categorization is also supported by discussion and agreement among all authors, together with external experts when needed. Yet, we agree that we might not be able to completely rule out subjectivity. 

\textbf{External validity} concerns the generalizability. Although we analyzed 654 issues from 10 popular RAG systems, which serve as good representations, our results may not fully generalize to all RAG systems, particularly proprietary systems. Our selection criteria favor popular and active open-source repositories with sufficiently rich issue histories. Therefore, the findings may not generalize to proprietary or long-tail RAG systems, newly emerging projects with fewer stars, or systems whose configuration problems are not publicly reported. Moreover, we did not explicitly select or analyze systems according to agentic or cyclic execution patterns. Caution is therefore required when extending our findings to newly emerging types of RAG systems involving planning, tool use, memory, iterative retrieval, or multi-agent orchestration.

\textbf{Construct validity} relates to the metrics/procedure used in the study. We choose Cohen’s $\kappa$ as it has been widely used for empirical studies in software engineering~\cite{9772253,cheriyan2021towards}; the same for Spearman correlation and Mann-Whitney U test. Further, we use elapsed calendar time as an observable proxy for resolving cost rather than a direct measurement of developers' active debugging effort. The interval may also include issue triage, waiting for maintainer responses, implementation scheduling, code review, and periods of inactivity. It might be affected by repository activity, maintainer availability, issue priority and severity, reporting quality, and project-specific development practices.




\section{Related Work}
\label{sec:related_work}


\textbf{Empirical studies on AI systems.}
A growing body of empirical research investigates reliability challenges in AI-driven software systems~\cite{DBLP:journals/corr/abs-2605-08305,8987482,10.1145/3530019.3530029,DBLP:conf/icse/Shao0S00025,xue2025characterization,chen2025empirical}.
{
Long and Chen~\cite{10.1145/3530019.3530029} investigate the reporting and resolution of performance and accuracy bugs in deep learning frameworks.
Shao et al.~\cite{DBLP:conf/icse/Shao0S00025} study LLM-integrated software applications and systematically analyzed their integration defects, while Xue et al.~\cite{xue2025characterization} examine the symptoms and root causes of bugs in LLM agent orchestration frameworks.
} 
{While these studies provide valuable empirical insights into the reliability and development challenges of AI systems, they primarily focus on deep learning frameworks and LLM-based applications, leaving the configuration of RAG systems largely unexplored.}
There are studies that systematically characterized the RAG landscape, detailing its architectural evolution and core paradigms~\cite{DBLP:journals/corr/abs-2312-10997, fan2024survey, arslan2024survey, oche2025systematic, csakar2025maximizing}. Among them, Gao et al.~\cite{DBLP:journals/corr/abs-2312-10997} provide a survey of RAG, categorizing its evolution into different types.
{Fan et al.~\cite{fan2024survey} systematically review mainstream relevant work by RAG architectures, training strategies, and application areas.
Şakar and Emekci~\cite{csakar2025maximizing} address the optimization of RAG processes by further exploring various methodologies.}
However, those works do not focus on real-world bug patterns or configuration-induced issues of RAG systems. Some other efforts are specific to RAG deployments~\cite{DBLP:conf/fllm/AgrawalKA024, DBLP:journals/corr/abs-2510-13975, DBLP:conf/cain/BarnettKTB024}.
{
Agrawal et al.~\cite{DBLP:conf/fllm/AgrawalKA024} identify failure points in knowledge graph-based RAG systems, while Leung et al.~\cite{DBLP:journals/corr/abs-2510-13975} develop a taxonomy of RAG errors and propose an automated evaluation approach.
Barnett et al.~\cite{DBLP:conf/cain/BarnettKTB024} identify seven failure points in RAG engineering through three case studies.
However, these studies primarily focus on RAG failure modes and output errors, without systematically investigating configuration-related issues and their relationships with the RAG pipeline.
}


{Unlike these studies, our work conducts a systematic empirical investigation of real-world configuration issues in RAG systems, characterizing their symptoms, root causes, and fix patterns. In particular, we examine how configuration issues manifest and originate across different RAG pipeline stages, revealing the discrepancies between symptom and root-cause locations and their implications for issue resolution which are previously unknown.}


\textbf{Empirical studies on configurable systems.}
Configuration-related errors have long been recognized as a key failure in general software systems.
Classic studies~\cite{10.1145/3243734.3243794, liu2024rethinking} analyze real-world misconfigurations to understand their root causes and impact on system availability.
Design-centric research~\cite{10.1145/2517349.2522727, xu2015hey} 
{explore the challenges of configuration design and usage, highlighting how inadequate configuration handling and excessive configuration options complicate system management and increase the likelihood of user errors.}
Performance-focused studies~\cite{han2016empirical, 10.1145/2345156.2254075, he2019tuning, he2020cp,10.1145/3803859,10.1145/3844143,DBLP:journals/tse/ChenGC25} characterize configuration-related performance bugs where internal code defects trigger performance degradation. {Other studies~\cite{10.1145/2345156.2254075, han2016empirical} empirically characterize real-world performance bugs, including the challenges of testing and debugging them in highly configurable systems. Subsequent research~\cite{he2019tuning, he2020cp} specifically investigates configuration-related performance bugs and develops detection techniques based on unexpected performance behaviors under different configuration settings.}




Being a new type of system, RAG systems introduce distinct configuration challenges that may limit the direct applicability of findings from traditional configurable systems, for two main reasons:
\begin{itemize}
    \item \textbf{Heterogeneous component integration.} RAG systems are integration-driven applications that depend on diverse external components, such as LLM APIs, embedding models, and vector databases. Their configurations must accommodate evolving interfaces, component dependencies, and compatibility requirements, which forms special categories for RAG systems such as \textit{inter-component implementation bug}. 
    
    
    {\item \textbf{Multi-stage pipeline.} RAG systems operate in their unique multi-stage pipelines, where configuration options influence inter-stage data flow and semantic processing logic. Configuration choices made at one stage can influence downstream data processing and system behavior, potentially causing failures to manifest at stages different from where their root causes originate. }
    
\end{itemize}

Those characteristics are what make our configuration study unique, as we unveil new, previously unknown findings/insights specifically on RAG systems.





\textbf{Configuration testing, tuning, and modeling.} 
{Configurable systems have been widely researched, including performance modeling~\cite{DBLP:journals/tse/GongCB25,siegmund2015performance,DBLP:conf/icse/XiangChen26} for learning the effects of configuration options and their interactions on system performance under varying conditions; configuration tuning~\cite{Chen_2021,chen2024mmometamultiobjectivizationsoftware,chen2025promisetune,10.1145/3832783.3834376,DBLP:conf/kbse/XiongC25,DBLP:conf/icse/Ye0L25} for efficiently searching for configurations that optimize system performance; and configuration testing~\cite{ma2025faster} for efficiently detecting configuration-related performance bugs.}
{Additionally, a major line of work focuses on the automated diagnosis of configuration errors and detection of performance bugs~\cite{10.5555/2486788.2486830,ma2025faster,han2018perflearner}, leveraging program analysis, historical bug reports, and testing prioritization techniques to identify and diagnose defects.
Other studies investigate automated configuration optimization and adaptation for RAG systems~\cite{barker2025faster,ray2025metis,11330856,DBLP:conf/ijcai/ChenC26}, aiming to improve answer quality, latency, and cost efficiency through parameter optimization, configuration selection, and workload-aware adaptation.}
Our study is complementary to those by revealing the unique characteristics of configuration issues for RAG systems.






\section{Conclusion}
\label{sec:conclusion}

In the paper, we conduct an empirical study consisting of 10 RAG systems and 654 configuration issues on GitHub. From this, we summarize a categorization that documents symptoms, root causes, relevant stages, and fix patterns for configuration issues of RAG systems. In particular, we reveal ten previously unknown findings. Drawing on those, we conclude several actionable insights for future configuration engineering of RAG systems. These results highlight that configuration reliability in RAG systems is not merely an option-selection concern, but a stage-dependent engineering challenge rooted in their integration complexity. We hope this work lays the foundation for more principled debugging and design practices of the next-generation RAG systems.





\bibliographystyle{ACM-Reference-Format}
\bibliography{ref}

\end{document}